\documentclass[letterpaper, 10 pt, conference]{ieeeconf}
\IEEEoverridecommandlockouts  %needed for \thanks
\usepackage{times}
\usepackage{graphicx}
\usepackage{subcaption}
\usepackage{url}
\usepackage{hhline}
\usepackage[font=small]{caption}
\usepackage{color}
\usepackage{amsmath, amsbsy} 
\usepackage{bm}
\usepackage{amssymb} 
\usepackage{gensymb}
\usepackage{multirow}
\usepackage{mathtools}
\usepackage{caption}
\usepackage{epsfig}
\usepackage{epstopdf}
\usepackage{bbold}
\usepackage{tikz}
\usepackage{relsize}
\usepackage[percent]{overpic}
\usepackage{xcolor}
\usepackage{siunitx}
\usepackage{tabularx}
\usepackage{algorithm,algcompatible}
\usepackage[noend]{algpseudocode}
\newcolumntype{Y}{>{\centering\arraybackslash}X}
\usetikzlibrary{shapes,arrows,decorations.pathmorphing,backgrounds,positioning,fit,petri}

\def\BState{\State\hskip-\ALG@thistlm}

\usepackage{booktabs}
\usepackage{xspace}
\usepackage{cite}
\usepackage{cleveref}

\newcommand\figref[1]{Fig.~{\ref{#1}}}

\newcommand\secref[1]{Sec.~{\ref{#1}}}
\newcommand\eqnref[1]{{Eq.~\eqref{#1}}}
\renewcommand\algref[1]{Alg.~{\ref{#1}}}

\title{\LARGE \bf 
Combined Task and Action Learning from Human Demonstrations \\for Mobile Manipulation Applications
}
\author{Tim Welschehold \and Nichola Abdo \and Christian Dornhege \and Wolfram Burgard 
  \thanks{All authors are with the Institute of Computer Science, University of Freiburg, Germany.
  This work has been supported by the Baden-W\"{u}rttemberg Stiftung and the German Research Foundation under research unit BU 865/8-1 (HYBRIS)}
}

\begin{document}

\maketitle

\begin{abstract}
  Learning from demonstrations is a promising paradigm for transferring knowledge to robots. However, learning mobile manipulation tasks directly from a human teacher is a complex problem as it requires learning models of both the overall task goal and of the underlying actions. Additionally, learning from a small number of demonstrations often introduces ambiguity with respect to the intention of the teacher, making it challenging to commit to one model for generalizing the task to new settings. In this paper, we present an approach to learning flexible mobile manipulation action models and task goal representations from teacher demonstrations. Our action models enable the robot to consider different likely outcomes of each action and to generate feasible trajectories for achieving them. Accordingly, we leverage a probabilistic framework based on Monte Carlo tree search to compute sequences of feasible actions imitating the teacher intention in new settings without requiring the teacher to specify an explicit goal state. We demonstrate the effectiveness of our approach in complex tasks carried out in real-world settings.
\end{abstract}
%\todo{
%\begin{itemize}
    %\item Sec IV A)
    %\begin{itemize}
        %\item Explain in detail $\omega_{(k,l)}$
        %\item Add something explaining the tackling of ambiguity in task goals, refer to fig 3 again (reviewer 2). (Did that in IV C) 1,2), TIM)
    %\end{itemize}
    %\item Sec IV B)
    %\begin{itemize}
        %\item An action $a$ defines which object should be moved and \textit{how} it should be moved, both on trajectory level and goal pose with respect to other objects in the scene. Each action is associated to one object $o_k$ and for each object $o_k$ there is only one associated action $a$. ()
        %\item Add details about the template definition, maybe example based on figure 3 (reviewer 1), (already updated figure 4 for better illustration, TIM) 
        %\item Set of actions $\mathcal{A}$ and set of templates for each action $\Gamma^a$ are discrete and finite (reviewer 1)
        %\item Remove the equation $p(\mathbf{s}_{t+1} | \mathbf{s}_t, a) = \sum_{\gamma \in \Gamma^a} p(\mathbf{s}_{t+1} | \mathbf{s}_t, a,\gamma) p(\gamma |\mathbf{s}_t, a) $. I don't think it helps understanding
    %\end{itemize}
    %\item Sec IV C)
    %\begin{itemize}
        %\item Explain generation of goal states finite set of discrete states generated clustering the demonstrations.. (Tim: added new subsusbsection \secref{sec:action_goals})
        %\item Describe leaf selection in more detail (Tim: added new subsusbsection)
        %\item Describe no-op action in text (and fig4, right branch) (TIM: done)
    %\end{itemize}
    %\item Video
    %\begin{itemize}
        %\item Add overview slide at beginning (reviewer 2)
        %\item Add sound?
    %\end{itemize}
%\end{itemize}
%}
\section{Introduction}

In order to operate intelligently in unstructured domestic environments, service robots should be able to handle a wide spectrum of situations and tasks, which makes it infeasible for an expert to pre-program a robot with sufficient knowledge to solve everyday tasks a-priori. In this context, learning from demonstrations is a promising paradigm as it allows non-expert users to instruct robots in an intuitive manner~\cite{12e2908dea004005a257707a1ad43237}.
%% new %specifically when considering arbitrary tasks that can be modeled geometrically like 
Ideally the learning is performed by observing a human teacher, as this avoids the restrictions posed by kinesthetic teaching in terms of limited motion of the robot base as well as requiring knowledge about the robot's kinematics. However, whereas this allows learning complex actions on a trajectory level, it introduces the necessity to map the observed human motion trajectories to the robot.%, taking into account its kinematics and grasping abilities. 

%\nichola{The order of arguments should be: 1) LfD is the way to go for teaching tasks. 2) Specifically, we consider tasks that can be modeled geometrically... Give a couple of examples like tidying up, opening cabinets, arranging objects, etc. 3) One way to do is kinesthetic, but this suffers from... Another is observing the teacher... *This allows learning complex actions on the trajectory level* but requires mapping ... 4) Another challenge is ambiguity (as is currently written but move problem of small number of demos to that paragraph). Learning the motion associated with the action is not enough as the robot also needs to learn the relevant spatial relations for the action, which is difficult when given a small number of examples.}
%However, learning by observing the teacher manipulating the objects in the scene is challenging since the robot has to map the observed trajectories to account for its own grasping capabilities and kinematics. 
%In addition, to achieve a practical system, robots should be able to learn tasks from a small number of teacher demonstrations. 
%In addition to learning the motion associated with an action, another challenge in this context is inferring the relevant spatial relations and geometric constraints from a small number of teacher demonstrations.

%is not enough as the robot also needs to learn the relevant spatial context of the action in an overall task, which is difficult given only a small number of examples.

In this work, we consider learning sequential mobile manipulation tasks that can be modeled geometrically based on the spatial relations between the involved objects, e.g., arranging objects on a table, tidying up, or operating doors. In this context, it is insufficient to learn the motion associated with executing each step of the task. Instead, the robot should also reason about the relevant spatial relations between the objects that it should achieve to solve the task and generalize the demonstrations in new settings. Without prior knowledge of the task, this introduces ambiguity with respect to the intention of the teacher and the goal of the task. For example, the pose of an object relative to the table might be more relevant for a task like setting the table than for a task like opening a box on the table. 
This makes it challenging to commit to one model for each action and for the overall goal of the task based on a few training examples. % , as this greatly limits the robot's ability to generalize in new settings. REMOVED, REPETITION
At the same time, without the user or an expert specifying an explicit goal to achieve in each situation, we cannot leverage existing task planning techniques to sequence the learned actions. 
%\nichola{Finally, in addition to learning the desired action goals }
\setlength{\tabcolsep}{1pt}
\begin{figure}[t]
	\centering
	\resizebox{0.75\columnwidth}{!}{%
  	\begin{tabular}{ccc}
  		\includegraphics[width=0.31\columnwidth,trim={6.0cm 0.9cm 7.5cm 0.65cm},clip]{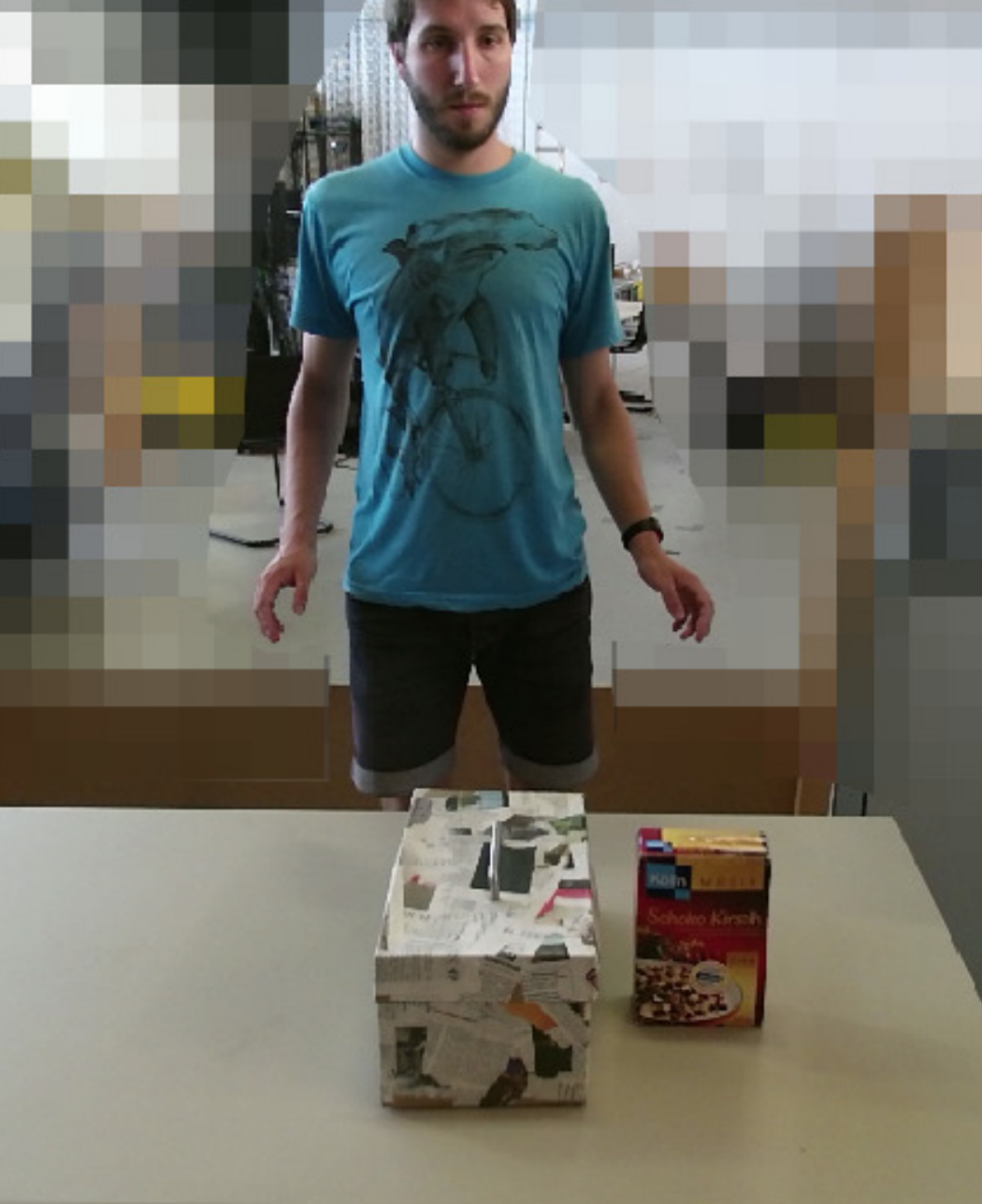} &
  		\includegraphics[width=0.31\columnwidth,trim={6.0cm 0.9cm 7.5cm 0.65cm},clip]{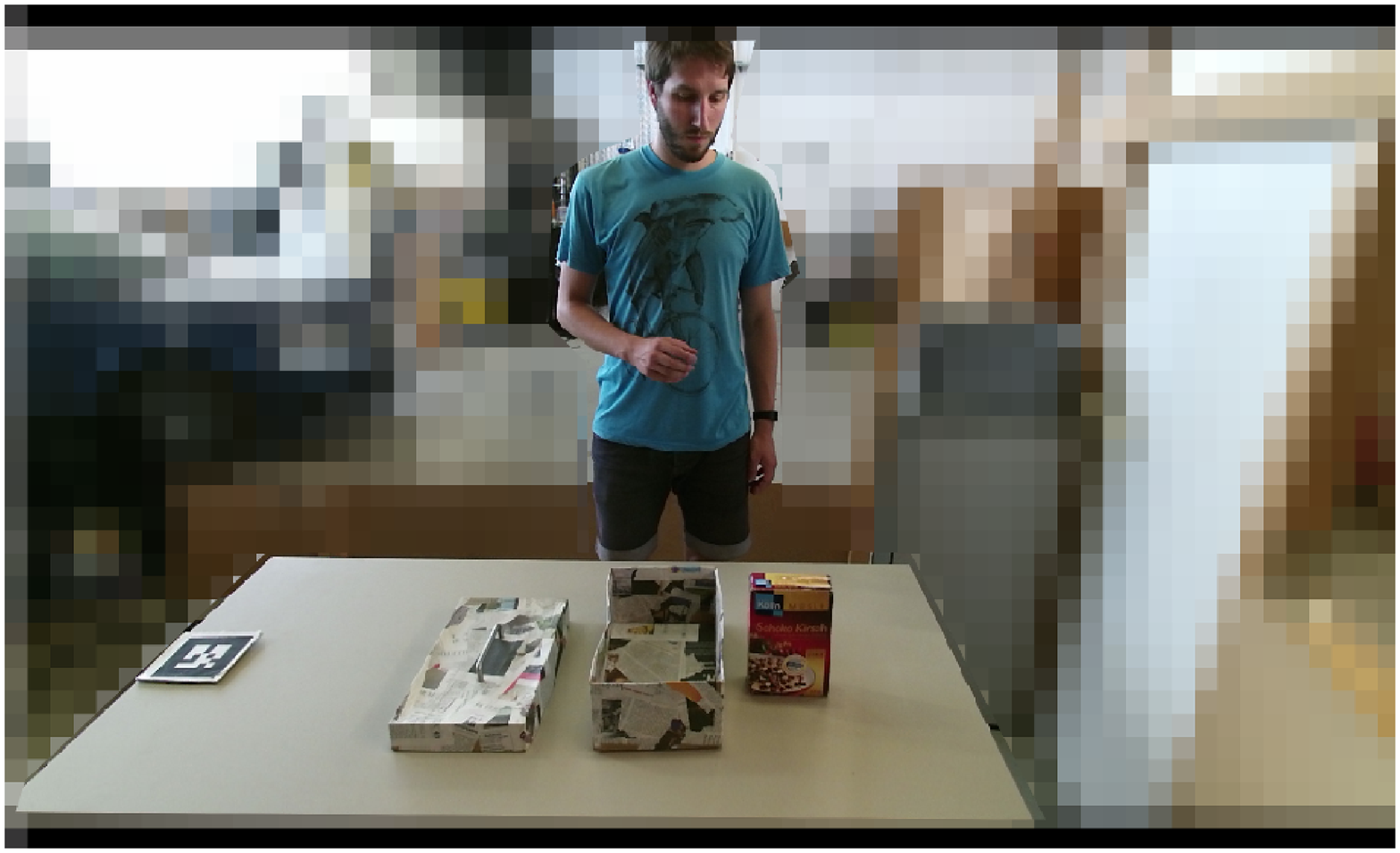} &
  		\includegraphics[width=0.31\columnwidth,trim={6.0cm 0.9cm 7.5cm 0.65cm},clip]{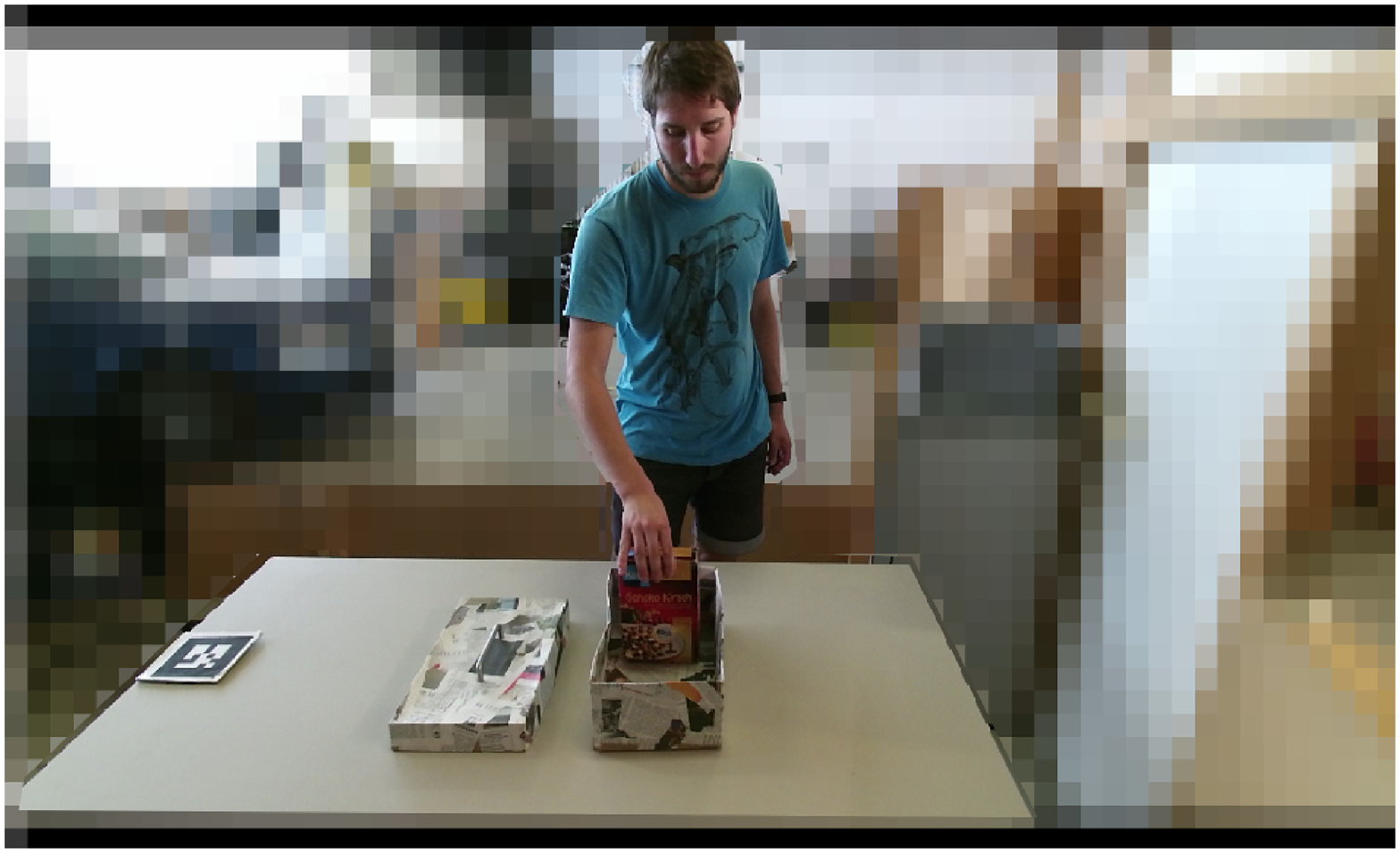}\\  		
 		\includegraphics[width=0.31\columnwidth,trim={17.3cm 4.5cm 14.7cm 15.5cm},clip]{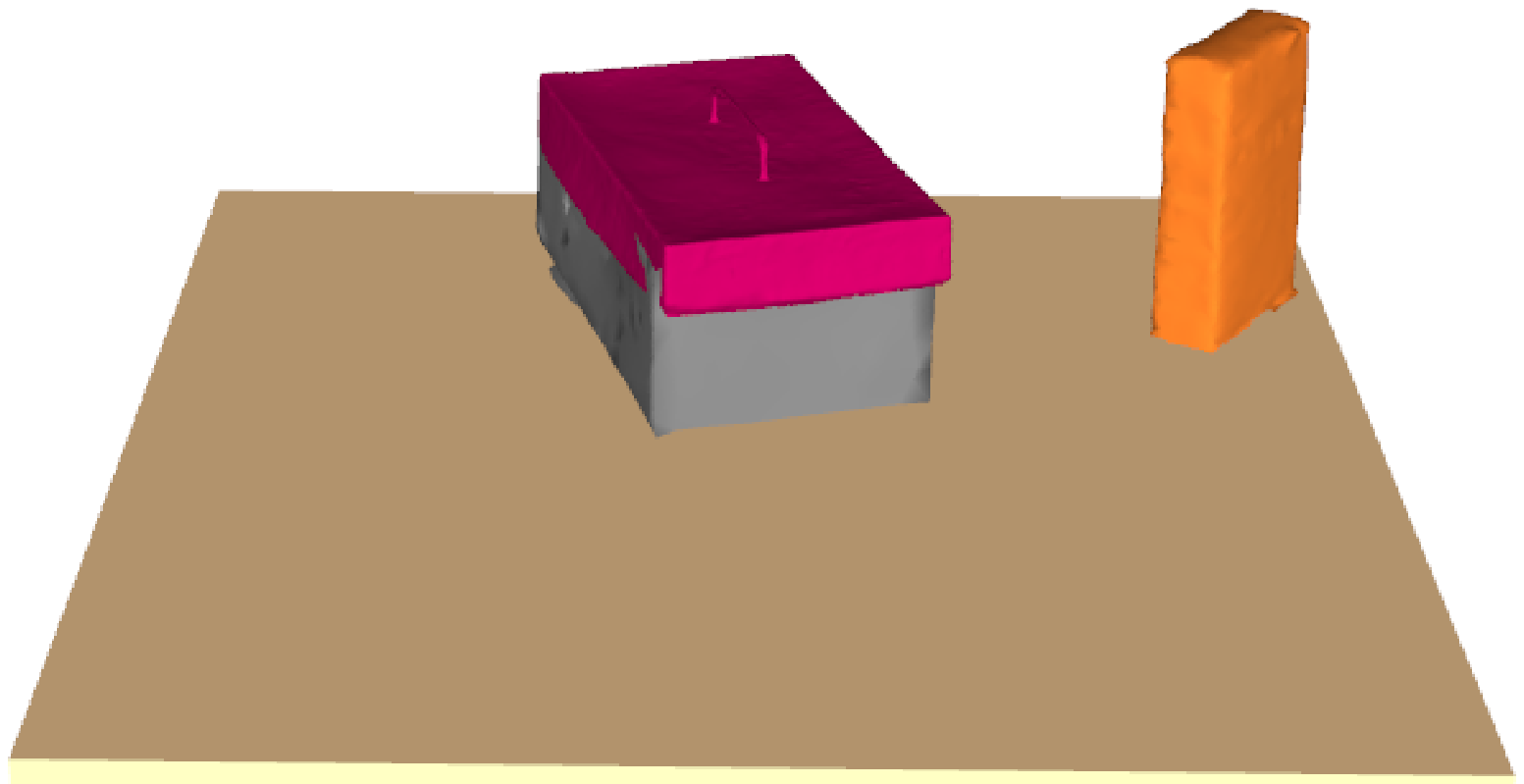} &
 		\includegraphics[width=0.31\columnwidth,trim={17.3cm 4.5cm 14.7cm 15.5cm},clip]{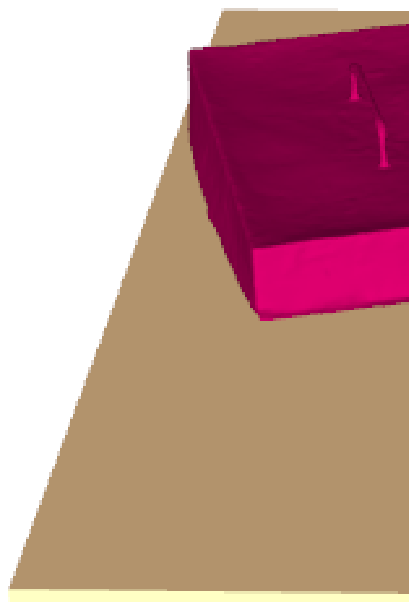} &
 		\includegraphics[width=0.31\columnwidth,trim={17.3cm 4.5cm 14.7cm 15.5cm},clip]{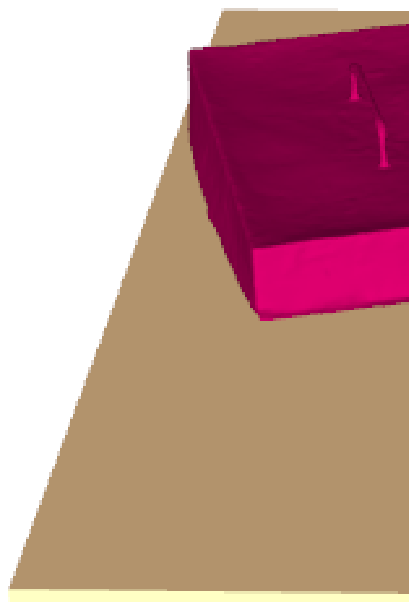}\\
 		\includegraphics[width=0.31\columnwidth,trim={12.9cm 4.5cm 14.7cm 1.4cm},clip]{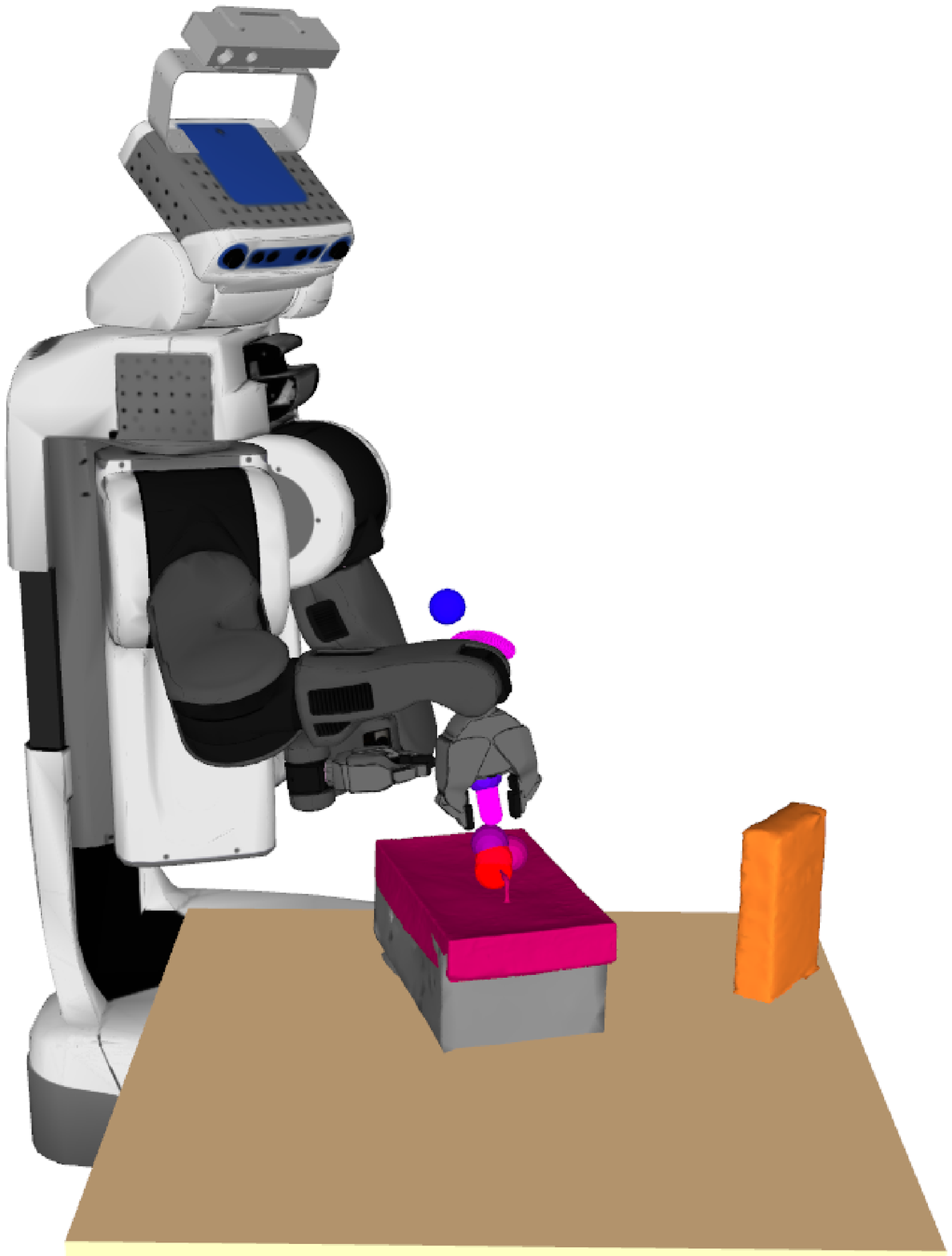} &
 		\includegraphics[width=0.31\columnwidth,trim={12.9cm 4.5cm 14.7cm 1.4cm},clip]{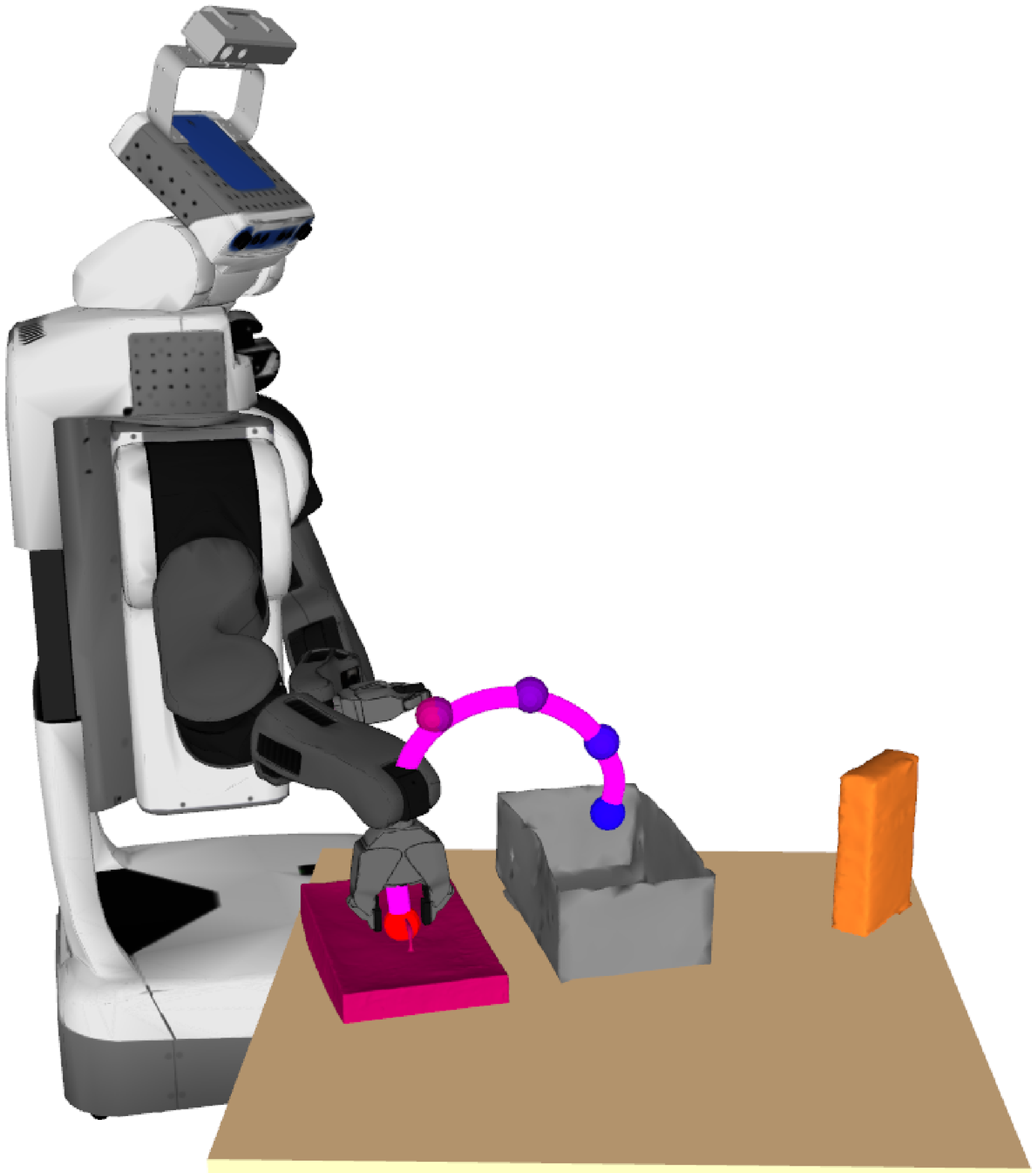} &
 		\includegraphics[width=0.31\columnwidth,trim={12.9cm 4.5cm 14.7cm 1.4cm},clip]{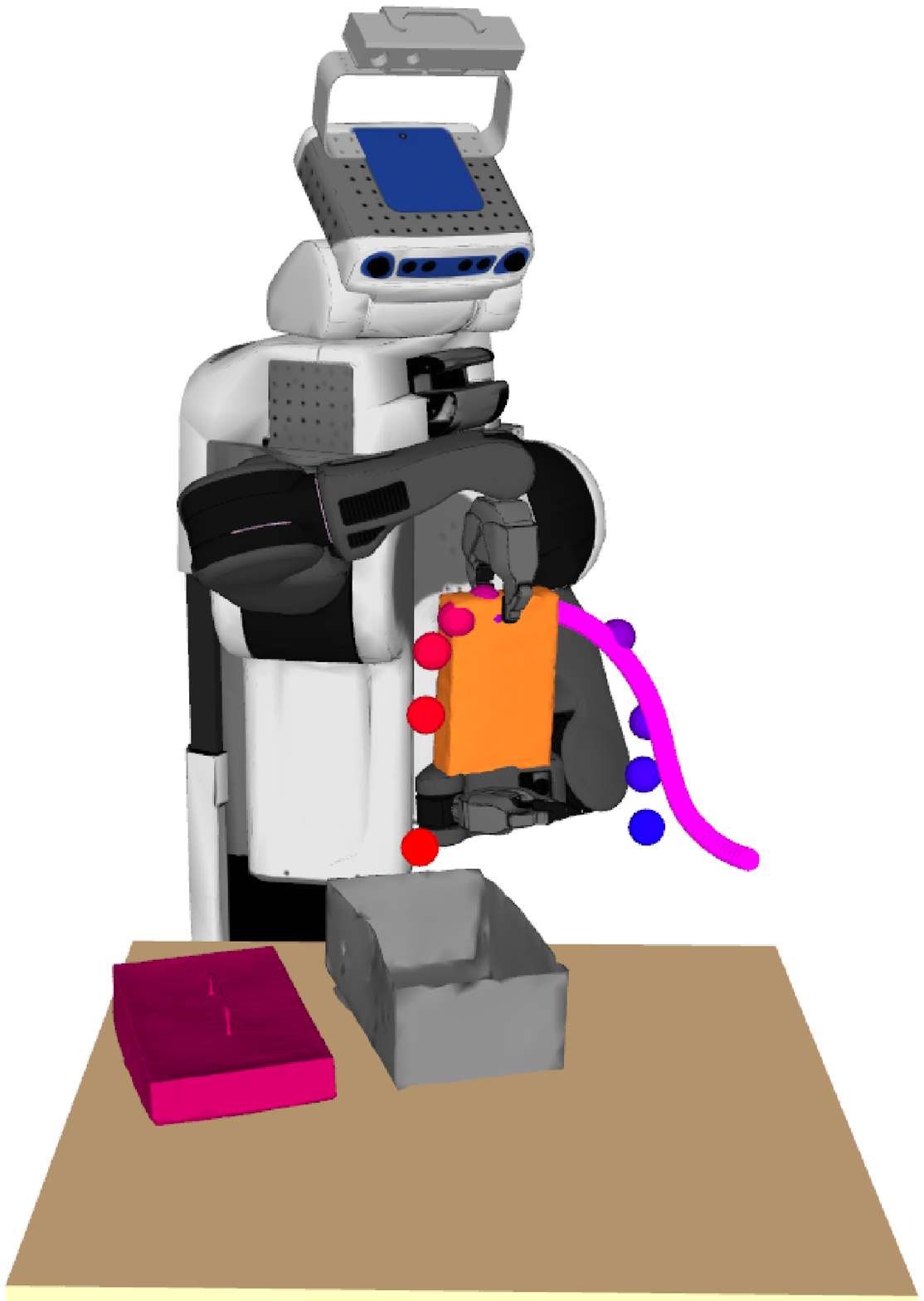}\\
	\end{tabular}
	}
	\caption{Our approach enables a robot to learn complex mobile manipulation tasks involving sequences of actions by observing teacher demonstrations (top row). By observing each step in the demonstrations, the robot is able to infer flexible models of the relevant spatial relations between the objects for each action and for the overall goal of the task
	(middle row)
	. At the same time, our approach enables the robot to learn action models on the trajectory level in order to execute the motions necessary to reproduce the task in new settings and without needing the teacher to specify a concrete goal of the task (bottom row).
	}
  	\label{fig:intro}
\end{figure}
\setlength{\tabcolsep}{6pt}

%In this work, we present a novel approach to learning mobile manipulation tasks that addresses the above challenges.
In our previous work~\cite{twelsche17iros}, we presented a novel approach to learning models of mobile manipulation actions that enables the robot to adapt complex human teacher demonstrations to account for the robot's kinematics and grasping capabilities. %, and without the restrictions associated with kinesthetic teaching. 
This allows the robot to learn motion models to reproduce the demonstrated actions by generating feasible trajectories.
%reproduce the motion needed to generalize the action in new settings. 
However, this approach considered one action at a time, and required the teacher to manually specify a reference frame (e.g., object on the table) to generate the motion. 
As opposed to that, in~\cite{abdo2017learning} we introduced \textit{teach-and-improvise}, a novel approach that builds on Monte Carlo tree search (MCTS) to enable the robot to compute sequences of demonstrated actions to achieve a goal state that aligns with the intention of the teacher. In contrast to existing planning techniques, our approach does not require committing to one model for each action or for the goal of the task. Instead, our algorithm disambiguates the demonstrations by considering several interpretations of the actions and the task with respect to the relevant spatial relations in a probabilistic framework. This allows the robot to improvise task solutions based on the starting state. One limitation of our work in~\cite{abdo2017learning}, however, was relying on existing motion planners to execute point-to-point actions for each step of the plan. This limited the imitation to tasks consisting of non-mobile and unconstrained actions. %\nichola{This limits the robot's ability to...}

In this work, we extend our previous works by combining the mobile manipulation
action models developed in~\cite{twelsche17iros} with the teach-and-improvise framework presented in~\cite{abdo2017learning}. Specifically, we present the following contributions: $\mathit{i)}$ our approach enables jointly learning models of a mobile manipulation task and its actions from a small number of markerless teacher demonstrations, and $\mathit{ii)}$ our approach enables the robot to imitate complex mobile manipulation tasks involving geometrically-constrained actions without requiring motion planners that depend on prior semantic knowledge of the task or an explicit goal representation. We evaluate our contributions thoroughly in real-world experiments with a PR2 robot. Note that an earlier version of this work with preliminary results was presented in~\cite{twelsche17lcriros}.

\section{Related Work}
\label{sec:rel_work}

\if 0
Learning from demos on the action/trajectory level. 
Learning from demos on the task level.
Geometric vs symbolic.
Using motion planners vs trusting trajectory from teacher.
Classical planning vs solving an optimization problem.
MCTS in general and heuristic-based.
\fi

%%%%%%%%%%%% Learning from demos on the action/trajectory level
In the field of learning actions on a trajectory level from demonstrations, different methods have been proposed over the last years~\cite{Pastor_ICRA_2009,Calinon12Hum}. In contrast to previous approaches we focus on adapting human demonstrations of mobile manipulation actions to the robot in terms of grasping capabilities and kinematic feasibility.
%\nichola{those are good but also old. Is there something newer for action trajectories you can cite? The most related thing to your previous papers}. 

%%%%%%%%%%%% Learning from demos on the task level.
Complementary to the learning of individual actions, there are also a number of approaches addressing imitation learning on a task level. Calinon~\emph{et al.}~\cite{ 5379592,4126276} extract spatial and temporal constraints and their relevance to each part of the task from demonstrations, while~\cite{5353894} automatically selects task spaces for modeling demonstrated motions. 
The work by Niekum et al. leverages a Bayesian nonparametric model to identify repeated structure in trajectories, and accordingly represents the high-level structure of the task using a finite state machine where each learned skill is encoded relative to the most consistent reference frame~\cite{doi:10.1177/0278364914554471}.
Such approaches rely on simple heuristics, statistical measures, or a teacher to determine relevant frames of reference for learning and executing motion primitives. In contrast to that, our method allows for considering several reference frames for each action in a Monte Carlo tree search framework.

%%%%%%%%%%%% Geometric vs symbolic.
Most current work on learning the integration  
%\todo{do you mean integration? -> yes, is embedding really wrong? --> No but it's a term usually reserved for learning embeddings in ML.} 
of action models in tasks builds on existing predicates to instantiate or extract symbolic or logical rules from teacher demonstrations such as the work by H\"ofer and Brock~\cite{Hfer2016CoupledLO}. Other approaches extract relevant preconditions and effects of actions from experience in a reinforcement learning domain~\cite{Konidaris:2014:CSR:2892753.2892821}.  Related to that, Pasula~\emph{et al.} introduced a probabilistic, relational planning rule representation that compactly models noisy, non-deterministic action effects~\cite{cbbe5d6ea9754310b0d0f99baa20ba96}. Recently, Paxton~\emph{et al.} proposed an approach that relies on a symbolic description of a task to perform sampling-based motion planning based on actions learned from expert demonstrations~\cite{paxton2016want}. Rather than extracting symbolic action representations for planning or relying on prior semantic knowledge such as a library of predicates or a planning domain, our approach addresses learning a continuous representation of actions and tasks from a small number of demonstrations. Accordingly, our action goal distributions encode multi-modal spatial relations between the objects.

%They learn symbolic action models from teacher demonstrations for action planning.\\

%\nichola{Move this part about Pasula's work (which is important) to earlier and closer to the other action works. Rephrase it a bit The focus should be that our representation is continuous. The connection should be that we have a probabilistic model of the action goals that captures the ambiguity in the demonstrations. This is related to their noisy non-deterministic rules etc. However, rather than learning these rules based on existing high-level relations, we do not assume such prior knowledge and we are able to model arbitrary, multi-modal spatial relations in a continuous space} 

%%%%%%%%%% MCTS in general and heuristic-based.
%\nichola{The MCTS paragraph should be as follows (please rephrase as I am only adding short hints now and not full sentences): Our approach to compute sequences of actions to solve a task builds on heuristic-based MCTS [survey paper by Brown, AND the work of Keller]. However, we extend the traditional tree structure of MCTS to enable the robot to consider multiple models for each action. This tackles the ambiguity in the teacher demonstrations and allows considering different interpretations bla bla in a probabilistic framework} 

To compute feasible sequences of actions when solving a task using our continuous representation, our algorithm builds on heuristic-based MCTS~\cite{97daaf7b774d446893dc3f5a1d945b49,Keller2013TrialBasedHT}. However, we extend the traditional tree structure of MCTS to leverage our flexible action models that encode several interpretations (possible reference frames) for each action.
%Combining Monte Carlo based search with heuristic methods was recently taken up in~\cite{Keller2013TrialBasedHT}. This is related to our approach as we initialize values for our search tree with a intention likelihood heuristic.

%%%%%%%%%%%% Classical planning vs solving an optimization problem.
Planning for robotics is typically addressed with task and motion planning techniques~\cite{dornhege09icaps,wolfe10icaps,petrick14icapsws
%,gregory12icaps,dornhege09ssrr,kaelbling13ijrr,gharbi2015
}. %\todo{Maybe remove some of these.} 
These integrate geometric reasoners into forward-chaining state-space search, Hierarchical Task Network planning, or knowledge-based planning. Such planning-based approaches require a description of the symbolic and geometric domain as well as a concrete goal state or formula. As opposed to this, our approach does not assume such prior semantic knowledge about the task as we build our models on a continuous representation modeling spatial relations between objects. Accordingly, rather than planning to achieve a pre-specified symbolic goal, we formulate task imitation as an optimization problem in which we maximize the likelihood that the final goal state aligns with the intention of the teacher.
% Speed-ups relevant
%%Such integrated task and motion planning systems suffer from the computational overhead of the geometric reasoners. This can be tackled, e.g., by only planning on a geometrical level to the point that the robot can execute an action~\cite{kaelbling11icra}.  The general problem to decide, when to backtrack in the geometric or symbolic world can be approached by interleaving both steps~\cite{burbridge13icaps}.
% Interleaved (like ours)
%%Here, failures to instantiate symbolic plans geometrically can also be used to compute failure reasons that give the symbolic planner better guidance~\cite{%srivastava13icapsws,srivastava14icra}.
% how we tackle this
% concrete models of (geometric) actions and goals to apply
Similar to task and motion planning methods, our approach integrates feasibility checks (e.g., collision checking) directly into the search. Additionally, we perform full feasibility checks when a plan is found to ensure it can be executed by the robot in practice. However, in contrast to the above planning approaches, we do not assume prior geometric models of actions (e.g., manipulating articulated doors) as we enable the robot to learn feasible trajectories from demonstrations that encode such constraints inherently.

%%%%%%%%%%%% Using motion planners vs trusting trajectory from teacher.
%\nichola{This one should be better integrated with some other paragraph:} 
Similar to us, Toussaint \emph{et al.}~formulate sequential manipulation tasks as an optimization problem without an explicit goal state~\cite{Toussaint:2015:LPO:2832415.2832517,18-toussaint-RSS}. However, they do not consider learning from teacher demonstrations, and leverage logic-geometric programming and continuous path optimization while considering symbolic, kinematic, and geometric constraints.
Furthermore, Medina \emph{et al.}~also tackle learning sequential tasks from demonstrations and focus on stable control policies and action transitions~\cite{pmlr-v78-medina17a}.

%Finally, Medina \emph{et al.}~also tackle learning sequential tasks from demonstrations and focus on learning time-independent control policies that result in stable behavior ~\cite{pmlr-v78-medina17a}.

%Medina \emph{et al.}~ also learn sequenced task from demonstrations. In contrast to us they use kinesthetic teaching and focus on the transition between individual actions~\cite{pmlr-v78-medina17a}.
%Unlike us, they do not address learning representations of the task and its actions from teacher demonstrations.
%Toussaint~\emph{et al.} present an approach for sequential manipulation planning including tool-use~\cite{18-toussaint-RSS}. Their work focuses and dynamic interaction with the environment using a predefined set of possible interactions. In contrast to our work they input knowledge about physical behavior of involved objects and they do not learn actions models nor task goals from demonstration. 

\section{Problem Statement}
\label{sec:problem}
\begin{figure}[t]
\centering
\resizebox{0.9\columnwidth}{!}{%

\begin{overpic}[width=1.0\linewidth,trim={0.0cm 1.0cm 0.0cm 4.5cm}]{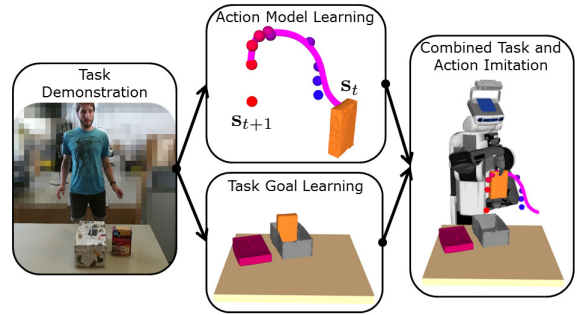}
  \put (58,38.0) {$\mathbf{s}_t$} 
  \put (39,32.0) {$\mathbf{s}_{t+1}$}
  \end{overpic}
  }
\caption{
Our approach uses the teacher demonstrations to learn models of the individual actions as well as the overall task goal. Each action manipulates one object to achieve desirable spatial relations between it and other objects in the scene. We learn an action goal distribution to sample likely goal states $\mathbf{s}_{t+1}$ to achieve when applying the action $a$ in $\mathbf{s}_{t}$. We leverage a Gaussian mixture model (colored spheres) to encode the motion of the action and generate feasible trajectories for executing it (magenta line). Our algorithm computes sequences of actions to imitate the task driven by maximizing the likelihood of the final task goal state achieved.
%From the human teacher task demonstrations we extract the information needed to learn both the models for the individual actions as well as the overall task intentions. The action models enable us to generate trajectories in the frame given by the handled object while the task goal imitation provides intermediate goal poses for all involved objects. The actions are encoded as \textit{Gaussian Mixture Models} (coloured spheres) and allow us to generate robot motion trajectories (magenta line). Combining both, we can than imitate the task without providing any explicit goals beforehand.
}
\label{fig:OverviewPic}
\end{figure}

We aim to enable a robot to learn a manipulation task from a small number of teacher demonstrations such that the robot can generalize the task in new settings.
We define a task as a tuple $\mathcal{T} = \left\langle \mathcal{O},\mathcal{A}, \Psi\right\rangle$ describing the involved objects $\mathcal{O}$, the set of applicable manipulation actions $\mathcal{A}$ and a function $\Psi$ modeling the teacher's intended task goal.
We represent the state $\mathbf{s}_t$ at time $t$ based on the 6-dof
poses of all objects involved in the task. 
%For this, we rely on existing perception solutions based on RGB-D observations of the objects and the teacher's hand and torso.
We use $^lT_k(t)$ to denote the pose of object $o_k$ relative to object $o_l$ at time $t$.
We consider learning tasks that involve several manipulation actions. 
%In this work, we do not address the problem of automatically segmenting the demonstrations into different actions. Instead, we rely on a heuristic to segment the demonstrations based on which objects are being manipulated by the teacher in each step. 
%\nichola{there needs to be a sentence here describing how we segment the demonstrations automatically based on which object is being manipulated since later we say we use the "segmented" demos.}
We do not explicitly address the problem of segmenting demonstrated trajectories in this work.
We assume that each manipulation action can be modeled based on three steps: reaching for and grasping an object, manipulating the object, and releasing the object. Accordingly, we automatically segment demonstrations based on co-occurring motion of objects and the teacher's hand. Beyond this we provide $3$D models of all objects involved but make no assumptions about the semantics of the task or its actions. Note that the teacher can choose to demonstrate the task using a different order of actions in each case. We assume that each task demonstration ends in a goal state that represents the teacher's intention for the task.
Accordingly, we aim to solve two joint problems, see \figref{fig:OverviewPic}:
First, given the segmented demonstrations, we aim to learn a set of actions $\mathcal{A}$ by adapting the observed trajectories such that the robot can reproduce them.
Second, given all task demonstrations, we aim to learn a model $\Psi(\mathbf{s})$ that captures how well a goal state $\mathbf{s}$ aligns with the intention of the teacher for the task. Accordingly, given a new starting state $\mathbf{s}_0$, we aim to compute a feasible plan consisting of actions $a_{0:T-1}$  from $\mathcal{A}$ and their subsequent goal states $\mathbf{s}_{1:T}$ such that the robot can achieve a final goal state $\mathbf{s}_T$ that maximizes $\Psi(\mathbf{s}_T)$. As there can be several ways of solving the same task,
%depending on $\mathbf{s}_0$,
we aim to do so without assuming knowledge of the goal state $\mathbf{s}_T$ or the number of steps $T$ needed to achieve it.

\section{Approach}

In this section, we present our approach for solving the problems in \secref{sec:problem}. We first describe how we model the intention likelihood $\Psi$ of the task in (\secref{sec:intention}). In \secref{sec:action_learning}, we describe our approach for learning action models that allow the robot to imitate the manipulation trajectories demonstrated by the teacher. In \secref{sec:tree_search}, we describe how we leverage those models in our teach-and-improvise framework to allow the robot to compute a feasible plan for solving the task starting in an arbitrary state.

\setlength{\tabcolsep}{1pt}
\begin{figure}[t!]
\centering
\resizebox{0.7\columnwidth}{!}{%
\begin{tabular}{cc}
	%\begin{minipage}{0.5\linewidth}
		%\includegraphics[width=0.85\linewidth,trim={1.0cm 5.0cm 1.0cm 2.0cm},clip]{./Bilder/a1} \vspace{.2cm} \\ \vspace{.2cm}
		%\includegraphics[width=0.85\linewidth,trim={1.0cm 5.0cm 1.0cm 2.0cm},clip]{./Bilder/a3} \\ 
		%\includegraphics[width=0.85\linewidth,trim={1.0cm 5.0cm 1.0cm 2.0cm},clip]{./Bilder/a10} 
	%\end{minipage} 

	\begin{minipage}{0.6\linewidth}
		\includegraphics[width=0.99\linewidth,trim={20.0cm 0.0cm 0.0cm 0.0cm},clip]{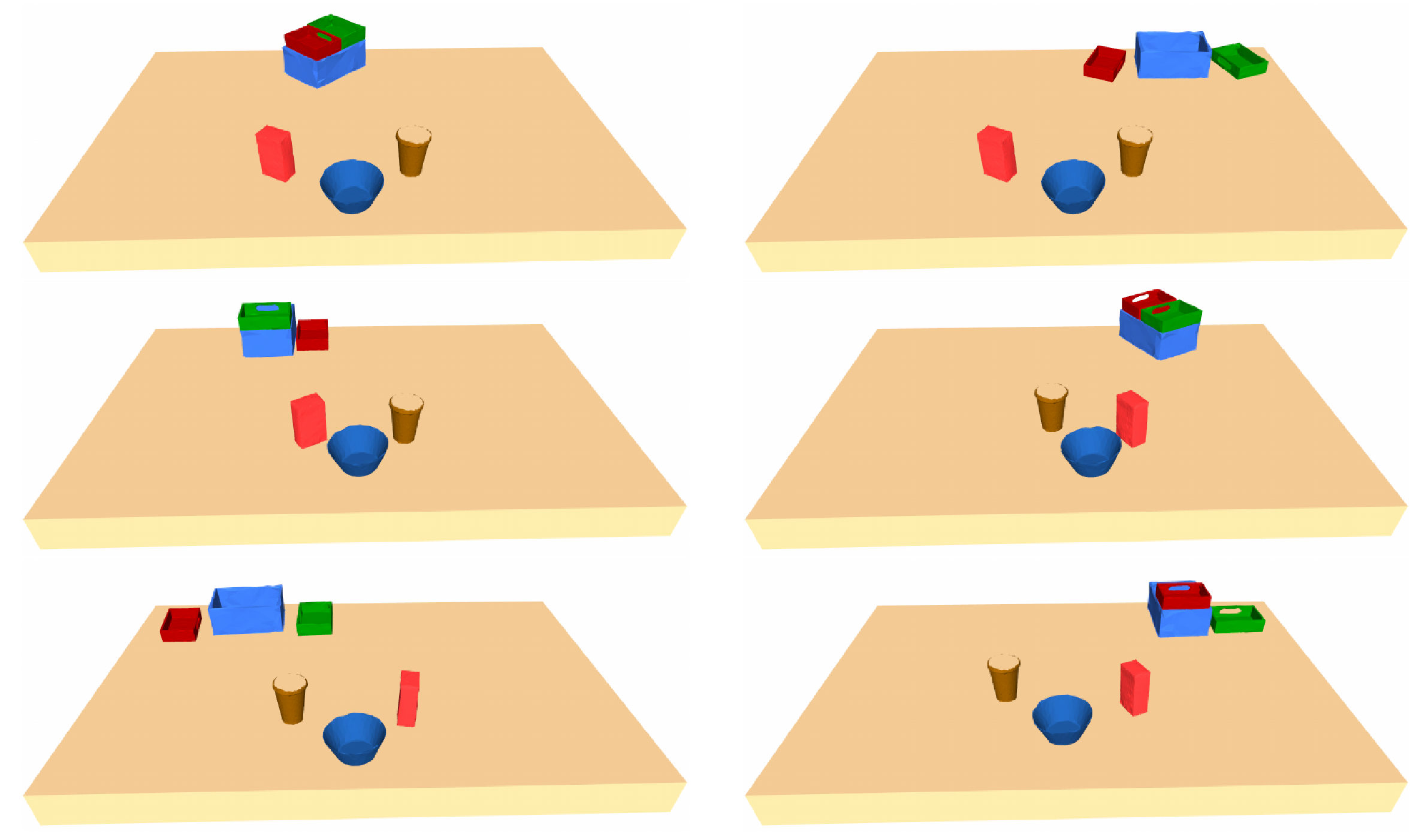} 
	\end{minipage}&
	\begin{minipage}{0.4\linewidth}
	\centering
		\includegraphics[width=0.8\linewidth,trim={00.0cm 0.0cm 0.0cm 0.0cm},clip]{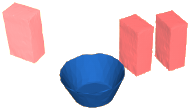}\\
		\includegraphics[width=0.8\linewidth,trim={00.0cm 0.0cm 0.0cm 0.0cm},clip,angle=5]{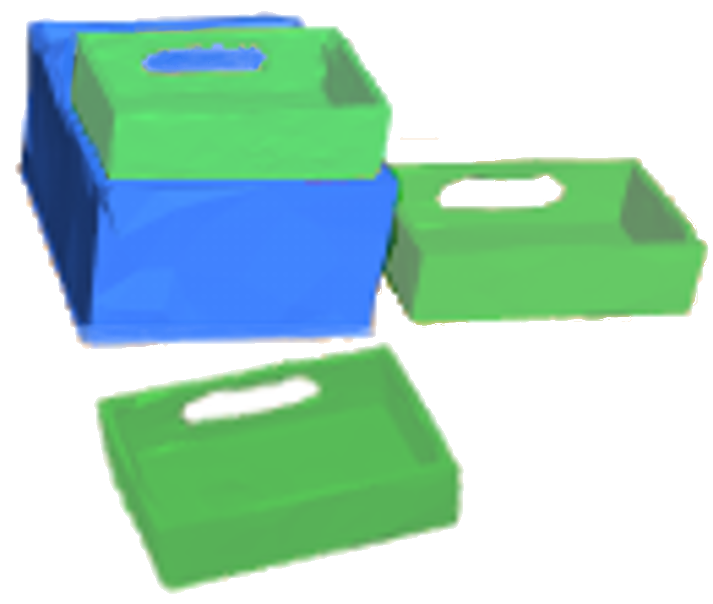}\\
		\includegraphics[width=1.4\linewidth,trim={10.0cm 5.0cm 0.0cm 0.0cm},clip]{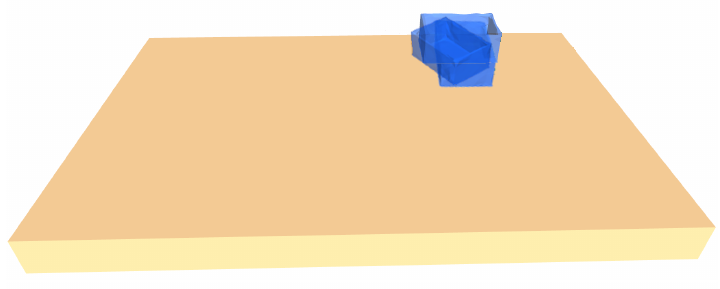}
	\end{minipage}
\end{tabular}
}
\caption{
The left column shows the final goal states for three different demonstrations of arranging objects on the table. The right column illustrates the inherent ambiguity in the demonstrations by showing the corresponding distributions of three pairwise relations from this task. We model these relations using multi-modal distributions we construct from the demonstrated poses. This flexible model allows us to encode multiple modes of solving the task by computing solutions that maximize the intention likelihood of the task.
%The left column shows the final states for three different demonstrations of arranging the objects on the table. The right column illustrates the distributions of pairwise relations and the potential ambiguity posed by different \textit{modes} of solving the task. The color intensity scales with the probability that the respective pose matches the intention of the teacher.
}
\label{fig:demonstrations}
\end{figure}
\setlength{\tabcolsep}{6pt}

\subsection{The Intention Likelihood of the Task}
\label{sec:intention}
One challenge of learning the demonstrated task is inferring the intention of the teacher with respect to the desired final goal state, see \figref{fig:demonstrations}. Rather than committing to one goal state $\mathbf{s}_T^*$ to reach for all cases, we introduce the intention likelihood function $\Psi(\mathbf{s}_T)$, which models the extent to which a goal state $\mathbf{s}_T$ aligns with the intention of the teacher for solving the task. As we do not assume to have any prior semantic knowledge about the task, we assume that $\Psi$ can be modeled based on the pairwise spatial relations between all pairs of objects in the task as follows:
\begin{equation}
    \Psi(\mathbf{s}_T) = \eta \sum_{o_k \in O(\mathbf{s}_T)}
    \sum_{\substack{o_l \in O(\mathbf{s}_T)\\ o_k \neq o_l }}
    \omega_{(k,l)} p({}^{l}\mathbf{T}_k(T)), %\frac{1}{N} \sum_{n=1}^{N} \mathbf{k}({}^{l}\mathbf{T}_k^{(n)}(T_n), {}^{l}\mathbf{T}_k)
\end{equation}
where $O(\mathbf{s}_T)$ is the set of objects in $\mathbf{s}_{T}$ and $p({}^{l}\mathbf{T}_k)$ is the likelihood that the pose of $o_k$ relative to $o_l$ in state $\mathbf{s}_T$ aligns with the corresponding training poses at the end of the teacher demonstrations. The weights $\omega_{(k,l)}$, normalized by $\eta$ capture the importance of the pairwise relation between $o_k$ and $o_l$ for the task, which we estimate from the consistency in the demonstrations % as described in~\cite{abdo2017learning}. This is based on the entropy of the relations, i.e., if all relations between $k$ and $l$ were similar in the demonstrations in contrast to always different.
%% new
by assuming that more relevant relations show a lower dispersion than less relevant ones. We define $\omega_{(k,l)} = \frac{1}{\epsilon_H+H_{(k,l)}}$,
%\begin{equation}
%    \omega_{(k,l)} = \frac{1}{\epsilon_H+H_{(k,l)}},
%\end{equation}
where $H_{(k,l)}$ is the entropy based on the object relations seen in the task demonstrations.  We ensure that all weights $\omega_{(k,l)}$ are finite and positive by setting $\epsilon_H = 0.01-\min(0,H_\textit{min})$, where $H_\textit{min}$ is the minimum entropy over all pairwise relations. We estimate the entropy of this distribution numerically by drawing samples from its $N$ modes as proposed by \cite{1055550}.
%% end new
%$\eta$ is a normalizer such that $1/(\eta)$ is equal to the sum of all weights. 
We aim to model the distributions $p({}^{l}\mathbf{T}_k)$ from a small
number of demonstrations and yet capture fine details of the teacher
intention. Accordingly, we adopt a data-driven approach based on
kernel density estimation to model these distributions from $N$
demonstrations:
\begin{equation} p({}^{l}\mathbf{T}_k) =
    \frac{1}{N} \sum_{n=1}^{N}
    k({}^{l}\mathbf{T}_k^{(n)}(T_n), {}^{l}\mathbf{T}_k),
\label{eqn:relation_distribution}
\end{equation} 
where $k(.,.)$ is the Gaussian kernel function capturing the
similarity between $^l\mathbf{T}_k$ and the corresponding relative pose
${}^{l}\mathbf{T}_k^{(n)}(T_n)$ at the end of the $n$-th demonstration.

%% new to address object being template for itself
%\nichola{The self-referencing thing is a bit hacky/confusing here. Do we need it here also or just in the actions?} Note that we also allow $k=l$, i.e., the relation of an object to itself. For this special case we use the relative position of the object at the beginning of the task with respect to its goal state instead of the relative state at the end of the demonstration. 

%% new to address the ambiguity concerns of reviewer 2
%- our model is flexible by considering a continuous distribution of goal states.
%- this addresses ambiguity in the demonstrations where it can be unclear what the intention of the teacher was, see fig.
%- this helps us avoid the need for committing to one goal state to always reproduce.
%- in section bla, we show how our mcts approach allows for solving the task to maximize this.
Our model of the intention likelihood is flexible and allows us to consider a continuous distribution of valid goal states for solving the task. This addresses the ambiguity in the demonstrations, where it can be unclear what the intention of the teacher is, see \figref{fig:demonstrations}. Rather than committing to a predefined goal state for reproducing the task using classical planning approaches, we show in \secref{sec:tree_search} how our MCTS-based framework enables the robot to \emph{improvise} feasible solutions that maximize $\Psi$.
%$\mathbf{s}_T$ that reproduces any of the demonstrated \textit{modes} to solve the task.
%\nichola{Strong opinion for changing fig 3: The left column of the figure is confusing and doesn't add much. I think the right column should become the left column (examples of goal states), and the right column should give ca 2 examples of pairwise relations from this task (or choose another task), as in fig 5.3 in thesis. This will address concerns about clarity about what the relations and equations mean intuitively.}

%As shown in \figref{fig:demonstrations}, the demonstrated goal states can vary among the demonstrations presenting different \textit{modes} of solving the task. Although this introduces ambiguity, the lower dispersion between relevant relations, e.g., between the blue bowl and the red cuboid compared to less relevant relations, e.g., between the blue bowl and the green container, will be reflected in the weights $\omega_{(k,l)}$ of the relations, giving higher intention likelihood $\Psi(\mathbf{s}_T)$ to a state.
%% end new

\subsection{Action Model Learning}
\label{sec:action_learning}
%\nichola{In my opinion, this section needs most of the clarification and added figures and then we can shrink some MCTS details. This is the section that can distinguish the paper from the thesis since it's the core of the added contribution, i.e., your work on learning the trajectories. In my opinion, it really needs a nice figure that explains the action model we use. It should show the object to be moved on the table with other objects next to it, and you can visualize the trajectory with a bunch of semi-transparent gripper poses that first reach the object, and then both the object and the gripper towards the goal. We can then easily illustrate what a template is in the figure based on the other objects. Essentially this is the expansion of one action step in the MCTS figure. I agree with the reviewer that this was very cryptic/unexplained in the current version. In my opinion, if we add this figure, we can drop figure 2 which currently isn't adding much. Regarding your work on learning the trajectory, we really need a bit more math and details there. Otherwise it's my work + 1 paragraph, and you are selling yourself short. Please add some in a way that the notation is still consistent. Do this first and then we worry about shrinking.}
Whereas $\Psi$ captures the intention of the teacher with respect to the final goal state of the task, we also aim to learn a library of actions $\mathcal{A}$ using the segmented demonstrations such that the robot can later sequence those actions to achieve a state maximizing $\Psi$. We assume that each action $a\in\mathcal{A}$ is associated with a unique object $o_k$ and consists of three steps: reaching for and grasping the object, manipulating the object, and releasing the object.
We therefore seek to learn a model that enables the robot to manipulate $o_k$ starting in a state $\mathbf{s}_t$ by defining: $1)$ how to sample a goal state $\mathbf{s}_{t+1}$ such that $o_k$ satisfies desirable \textit{spatial relations} between $o_k$ and other objects in the scene after applying the action, and $2)$ how to generate a \emph{feasible trajectory} for moving $o_k$ to achieve $\mathbf{s}_{t+1}$ starting from $\mathbf{s}_{t}$.
%Accordingly we model each action both on trajectory level and regarding goal pose with respect to other objects in the scene.
%Each action is unique with respect to which object $o_k$ is being manipulated and for each object $o_k$ there is only one associated action $a$.
%We model each action on two levels:
%First, based on the spatial relations between the objects before and after applying the action. 
\subsubsection{Action Templates and Goal Distributions}
\label{sec:action_templates}
%- We need a distribution to sample goal states.
%- We build this on the spatial relations between the object being moved and the other objects.
%- We model this based on the training poses observed in demos.
%- For each pair of objects, we consider a distribution as in eq.
%- Ambiguity here again. The teacher can be moving relative to any object.
%- Sampling from the whole thing is expensive and will lead to needlessly exploring irrelevant regions of the state space.
%- We tackle this with templates.
We aim to model a goal distribution $p(\mathbf{s}_{t+1} \mid \mathbf{s}_t, a)$ that captures the likelihood to achieve state $\mathbf{s}_{t+1}$ when applying action $a$ to move $o_k$ in state $\mathbf{s}_t$. We assume that the intention of the teacher is to achieve desirable spatial relations between $o_k$ and other objects in the scene. Analogous to the intention likelihood of the whole task $\Psi$, we adopt a non-parametric mixture model for $p(\mathbf{s}_{t+1} \mid \mathbf{s}_t, a)$ based on the training poses before and after each teacher demonstration of the action. Specifically, we consider the poses of $o_k$ relative to all other objects. This enables us to tackle the ambiguity in the demonstrations and not commit to a single model of the action (e.g., always moving $o_k$ relative to $o_l$).
However, we also aim for an efficient way to sample from this distribution when searching for a solution to the task (\secref{sec:tree_search}). Therefore, we propose to decompose it by considering several action \textit{templates} $\Gamma^{a}$, where each template $\gamma\in\Gamma^a$ conditions moving $o_k$ relative to only one other object in the scene, i.e., $p(\mathbf{s}_{t+1} \mid \mathbf{s}_t, a) = \sum_{\gamma \in \Gamma^a} p(\mathbf{s}_{t+1} \mid \mathbf{s}_t, a,\gamma)\ p(\gamma \mid \mathbf{s}_t, a) $. Without any prior knowledge about the action, we assume $p(\gamma \mid \mathbf{s}_t, a) = \frac{1}{|\Gamma^a|}$ for all $\gamma \in \Gamma^a$. We construct a mixture distribution $p(\mathbf{s}_{t+1} \mid \mathbf{s}_t, a,\gamma)$ for each template (analogous to \eqnref{eqn:relation_distribution}) based on the poses ${}^{l}\mathbf{T}_k$ of $o_k$ relative to one object $o_l$ in $\mathbf{s}_{t+1}$ as seen in the action demonstrations.  This allows the robot to explore a different subset of spatial relations at a time depending on the situation. For example, it might be more beneficial to move a cup to achieve a desired pose relative to a bowl ($\gamma_1$) than to the door ($\gamma_2$). Note that $\Gamma^a$ includes the special template of moving an object relative to itself (e.g., the cabinet door in the second row of \figref{fig:combined_learning}). In \secref{sec:node_expansion}, we describe how our algorithm samples from this distribution when solving the task.
\subsubsection{Learning the Action Trajectory}
\label{sec:action_traj}
In addition to reasoning about a likely goal state $\mathbf{s}_{t+1}$ of the action, we aim to learn the motion associated with executing the action in state $\mathbf{s}_t$ to achieve $\mathbf{s}_{t+1}$ on the trajectory level. For this, we adopt the approach we introduced in~\cite{twelsche17iros} for learning mobile manipulation actions directly from human demonstrations. This allows the robot to adapt the demonstrations to its capabilities to achieve successful grasps of the manipulated object while considering kinematic constraints between the robot's mobile base and end-effector. We formulate this as a graph optimization problem that incorporates these constraints and generates robot-suited trajectories for grasping, moving and releasing the object by following the human demonstrations as close as possible while deviating as necessary to meet the robot's capabilities. We use the adapted demonstrated trajectories to learn models for a combined motion of the robot's base and end-effector described by their pose, velocity, and acceleration. The motion is assumed to be driven by a second order differential system encoded as a \textit{Gaussian Mixture Model}. % \cite{Calinon12Hum}. 
\figref{fig:OverviewPic} and \figref{fig:combined_learning} include visualizations of the learned time dependent models.% \nichola{this is where one would reference the new action fig I suggested}

For details on the graph structure and the implementation of our mobile manipulation action models, we refer to our work in ~\cite{twelsche17iros,twelsche18iros}. The learned models can generate trajectories to grasp, move and release the action-relevant object $o_k$ on demand. However, one limitation of our prior work is assuming that the reference frame for the motion and accordingly, the goal pose for the manipulation is provided by the teacher. In the next section, we discuss how this work addresses this issue by enabling the robot to automatically select different reference frames based on the action templates $\Gamma^a$ and sample goal states from the corresponding distributions (\secref{sec:action_templates}) while computing a task plan.
%We use the adapted demonstrated trajectories to learn models for a combined motion of the robot's base and end-effector described by their pose $\boldsymbol{\xi}$, velocity $\dot{\boldsymbol{\xi}}$ and acceleration $\ddot{\boldsymbol{\xi}}$. The motion is assumed to be driven by a second order differential system \cite{Calinon12Hum}:
%\begin{equation}
%	\label{eq:dynsys1}
%	\ddot{\boldsymbol{\xi}} = K_p(\hat{\boldsymbol{\eta}}-\boldsymbol{\xi}) - K_\nu \cdot %\dot{\boldsymbol{\xi}}.
%\end{equation}
%Where $K_p$ and $K_\nu$ are fixed parameters simulating the system's stiffness and damping. $\hat{\boldsymbol{\eta}}$ is the path of a virtual attractor guiding the motion. It is encoded as a \textit{Gaussian Mixture Model} based on the adapted demonstrations. Visualizations of the learned time dependent models are shown in \figref{fig:OverviewPic} and \figref{fig:combined_learning}. For details on the graph structure and the implementation of the action models we refer to our recent work~\cite{twelsche17iros}. The learned action models are flexible with respect to robot base motion and positioning~\cite{twelsche18iros}. 
%Each action stores the relative grasp pose to its related object and can generate trajectories to grasp, move and release it on demand. In order to be applied the actions require a corresponding reference frame $\gamma$ and a goal pose. In the next section, we discuss how our approach enables automatically selecting different reference frames $\gamma$ and propose new object poses while computing sequences of actions to solve the task.

\begin{algorithm}
\caption{The proposed teach-and-improvise algorithm based on the
    initial state $\mathbf{s}_0$ and intention likelihood $\Psi$ with
    feasibility checking for the applied action models $\mathcal{A}$.
%We execute the search for K iterations or until the root node has been solved, i.e., no more node expansion is possible
} 
\label{alg:teach_and_improvise}
\begin{algorithmic}[1]
\Procedure{SolveTask}{$\Psi$, $\mathcal{A}$, $\mathbf{s}_0$, $K$}
%\State // Create root node
%\State $\textit{prob} \gets 1$ \COMMENT{initial state probability}
\State $n_a \gets \textsc{CreateRootNode}(\Psi, \mathbf{s}_0)$
\State // Run $k$ iterations or root is solved ($\rho(n_a) = \textit{true}$)
\State $k \gets 0$
\While{$k<K$ \textbf{and} $\rho(n_a) = \textit{false}$} 
 \State $n^*_a \gets \textsc{SelectLeafNode}(n_a)$ %\COMMENT{select leaf}
 \State $\textsc{ExpandNode}(n^*_a, \Psi, \mathcal{A})$  %\COMMENT{Expand node}
 \State $\textsc{UpdateValues}(n^*_a)$  %\COMMENT{backup results to the root}
 \State $k \gets k+1$
\EndWhile
\State // Find best plan with feasible actions
\State $\textit{found} \gets \textit{false}$
\While{\textbf{not} $\textit{found}$}
 \State $\textit{bestPlan} \gets \textsc{RecommendBestPlan}(n_a)$
 \State $\textit{found} \gets \textsc{CheckFeasibility}(\textit{bestPlan})$
 \If{\textbf{not} $\textit{found}$} 
   $\textsc{UpdateValues}(n^*_a)$ 
 \Else{} 
  \Return $\textit{bestPlan}$
 \EndIf
\EndWhile

\EndProcedure
\end{algorithmic}
\end{algorithm}

\subsection{Solving the Task through Search-Based Optimization}
\label{sec:tree_search}

In this section, we describe our approach for solving a task given the
intention likelihood and action models learned above. For this, we
build on the teach-and-improvise framework we introduced
in~\cite{abdo2017learning}. Rather than planning to achieve
a predefined goal state, we formulate solving the task as the problem
of computing a feasible plan that maximizes the intention likelihood, i.e.,
\begin{equation}
	\max_{\substack{a_{0:T-1}\\ \gamma_{0:T-1}\\ \mathbf{s}_{1:T}}} 
		\Psi(\mathbf{s}_T) - \sum_{t=0}^{T-1}\mathit{cost}(a_t).
\end{equation}
The plan consists of a sequence of actions $a_{0:T-1}$, the corresponding templates $\gamma_{0:T-1}$ used to apply each action, as well as the intermediate goal states $\mathbf{s}_{1:T}$ resulting from applying each action. We additionally pose the constraint that the computed plan must be executable by the robot. 
We do not assume the best number of actions $T$ to solve the task is known beforehand. Accordingly, we use a constant $\mathit{cost}(\cdot)$ for each action to favor more efficient plans. This formulation enables the robot to use the learned models and “improvise” a suitable goal state $\mathbf{s}_T$ for solving the task depending on the starting state $\mathbf{s}_0$.
The optimization involves discrete (actions and templates) and continuous (goal states) variables, as well as a non-convex objective function $\Psi$ and feasibility constraints. To tackle these challenges, our algorithm builds  on  Monte  Carlo  tree  search as described next.
%Due to the non-convexity of $\Psi$, the ambiguity in the action and task goals, and the feasibility constraints, our algorithm builds on Monte Carlo tree search to solve this problem as described next. %\nichola{add sth about optimizing over discrete and continuous variables at the same time}

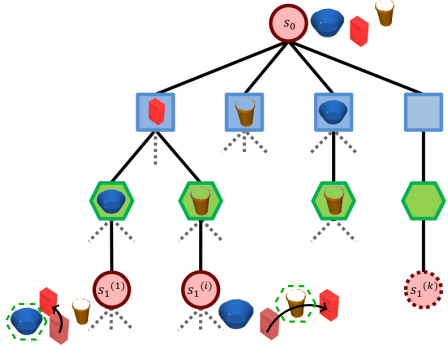
\begin{figure}
    \centering
    \resizebox{0.7\columnwidth}{!}{%
    \begin{tikzpicture}[]
    \node[anchor=north west] (image) at (0,0) 
  		{\includegraphics[width=1.0\linewidth,trim={1.0cm 24.2cm 4.5cm 0.2cm},clip]{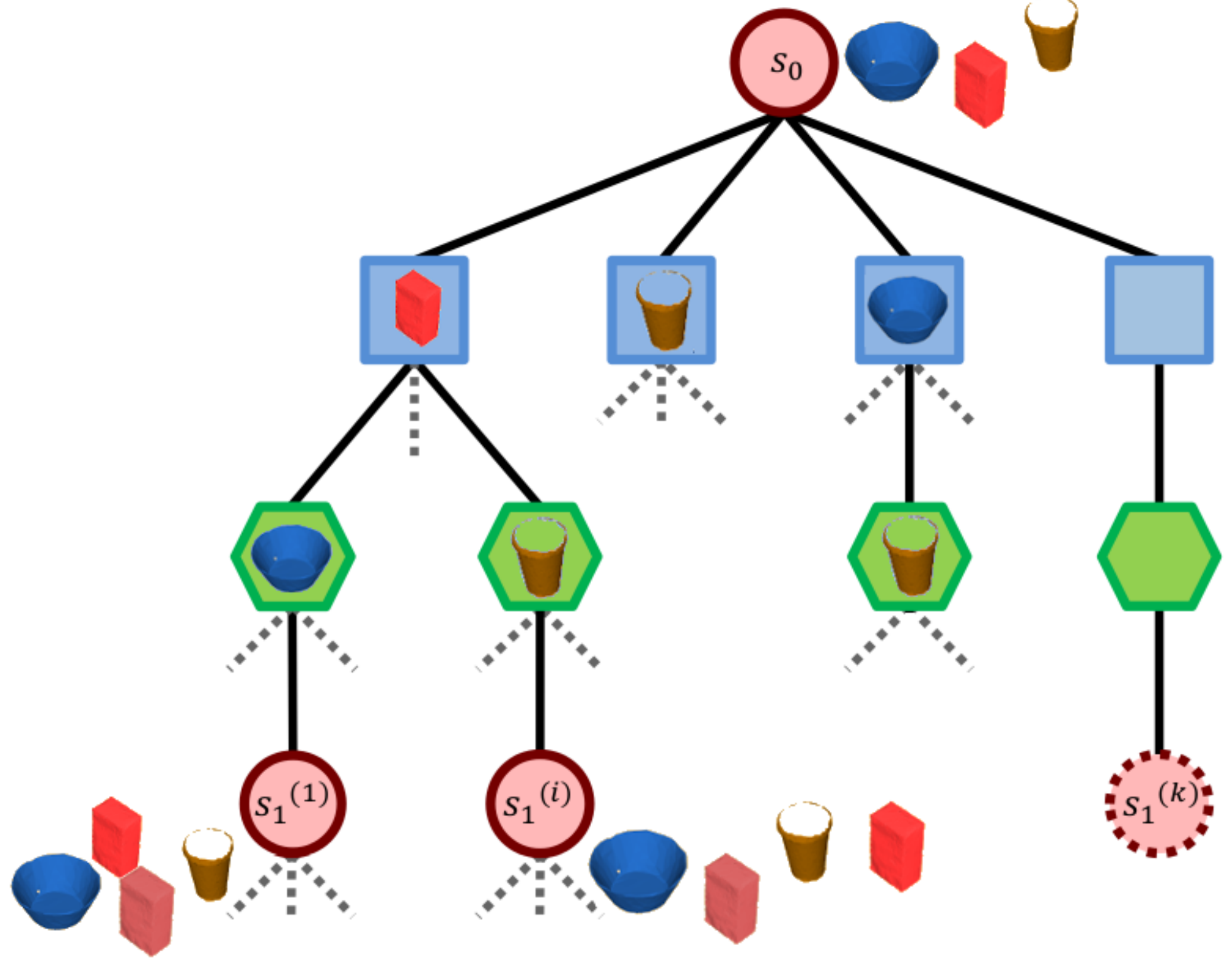}};
  		\draw[arrows=->,bend angle=45, bend right,line width=1.2pt](1.15,-6.6) to (1.0,-6.1) ;
  		\draw[arrows=->,bend angle=45, bend left,line width=1.2pt](5.2,-6.6) to (6.38,-6.15);
  		\newdimen\R
        \R=0.4cm
        \draw[yshift=-6.39cm,xshift=0.5cm,draw=green!70!black,line width=1.2pt,dashed] 
        (0:\R) \foreach \x in {60,120,...,360} {  -- (\x:\R) };
        \draw[yshift=-6.04cm,xshift=5.75cm,draw=green!70!black,line width=1.2pt,dashed] 
        (0:\R) \foreach \x in {60,120,...,360} {  -- (\x:\R) };
    \end{tikzpicture}}
    \caption{
    An illustration of our Monte Carlo tree search structure. The red circles represent action-selection nodes, which are used to reason about which action to apply from state $\mathbf{s}$, i.e., which object should be moved. Template selection nodes are depicted as blue squares and are used to select templates for the corresponding actions, i.e. given the selected action (or object to move, illustrated inside the squares) they select which object should be used as a reference frame for the motion. The green hexagons depict goal-selection nodes used for selecting new goal poses, i.e., given the state $\mathbf{s}_t$ of their grand-parent, the selected action $a$ of their parent, and the selected reference frame $\gamma$ (illustrated inside the hexagons), they sample new possible goal states $\mathbf{s}_{t+1}$ from the learned action goal distribution. The branch on the right depicts the special no-op action used to ensure termination.
    %Proposed structure for the Monte Carlo Tree Search. The red circles define action-selection nodes, which are used to reason about which action to apply from state $\mathbf{s}$, i.e., which object should be moved. Template selection nodes are depicted as blue squares and are used to select templates for the corresponding actions, i.e. given the selected action (or object to move, illustrated inside the squares) they select which stationary object should be used as reference for moving the object. The green hexagons describe goal-selection nodes used for selecting new goal poses, i.e., given the state $\mathbf{s}_t$ from their grand-parent, the selected action $a$ from their parent and the selected object as reference frame $\gamma$ (illustrated inside the hexagons) they sample new possible goal states $\mathbf{s}_{t+1}$ . The branch on the right describes the special no-op action used to ensure termination.
    }
\label{fig:tree}
\end{figure}

\subsubsection{Tree Structure}
\label{sec:tree_structure}
Classical MCTS iteratively grows a search tree starting from $\mathbf{s}_0$ to approximate the returns of states and actions using Monte Carlo simulations while trading off exploration and exploitation, see~\cite{97daaf7b774d446893dc3f5a1d945b49}. In the context of Markov Decision Processes, MCTS typically considers two types of tree nodes: decision and chance nodes. Decision nodes select actions to apply in their associated states. Chance nodes reason about potential outcomes of the action, usually reflecting the stochasticity of the domain. In our context of imitating a demonstrated task, we assume that the stochasticity stems from the ambiguity in the goals of the actions and the task. Given our approach to leverage several interpretations (templates) of each action as in \secref{sec:action_templates}, we propose a tree structure involving three types of nodes, see \figref{fig:tree}. \textit{Action-selection nodes} reason about which action $a \in \mathcal{A}$ to apply in a certain state. Subsequently, \textit{template-selection nodes} reason about which template $\gamma\in\Gamma^a$ to use for applying the action $a$ of their parent nodes. Each child of a template-selection node considers one template (reference frame) for the action, and is referred to as a \textit{goal-selection node}. Each such node is responsible for sampling goal states (new action-selection nodes) of the action from its goal distribution conditioned on the template and starting state as described in \secref{sec:action_learning}. This structure allows us to tackle the dimensionality of the problem and disambiguate the demonstrations by efficiently searching through the space of potential action interpretations and their intended goals.
%% new -- no-op action 
As the optimal number of steps for solving the task is unknown beforehand, we address the stopping problem by incorporating a special no-op action $a^0$ in the set $\mathcal{A}$, which does not change the current state and leads to a goal state that cannot be further expanded.

\subsubsection{Anytime Tree Search}
\label{sec:anytime_search}
We adopt an anytime search approach with a budget of $K$ iterations. \algref{alg:teach_and_improvise} summarizes our algorithm.
%during which we update the tree with the best node values found thus far. 
%Each iteration consists of three main steps.
After initializing the tree with the starting state as root (line 2) we iteratively select a promising leaf action-selection node (line 6) and expand it by applying all possible actions from the set of actions $\mathcal{A}$ in its state $\mathbf{s}_t$ (line 7). For each action, this adds subsequent template-selection, goal-selection, and action-selection children nodes (new leaves representing the resulting goal states $\mathbf{s}_{t+1}$).
Each node is associated with a value continuously updated to reflect its utility for the solution. Therefore, each iteration initializes the values of newly-added nodes and backpropagates them to the root (line 8). Those values guide the selection of new leaf nodes in subsequent iterations based on exploration-exploitation considerations.
% (\secref{sec:leaf_node})
%~\cite{abdo2017learning}. 
%% new -- no-op action termination
We repeat this process for $K$ iterations or until no leaf can be further expanded.
%We repeat this process until the budget of $K$ iterations is used up or no leaf can be further expanded.
Due to space limitations, we now briefly describe the idea behind each of these steps, and refer the reader to our work in~\cite{abdo2017learning} for all variations and implementation details of the algorithm.
%% new
%Leafs resulting from applying the no-op action $a^0$ are never expanded an thus serve as termination criteria by indicating whether applying any further action will decrease the intention likelihood of the generated scene. 

\paragraph{Node Selection}
\label{sec:node_selection}
Selecting a promising node to expand (line 6) involves traversing the tree from the root by sampling one child of each node iteratively until a leaf node is reached. For action-selection and template-selection nodes, rather than greedily selecting the child with the highest value, in this work we employ a Boltzmann Exploration strategy to trade off exploration with exploitation. This initially encourages a more uniform sampling of children, and gradually converges to greedily selecting the child with the highest value as the visit count of the parent increases.
%we adopt strategies that trade-off exploration with exploitation. In this work, we rely on the \textcolor{red}{BLA strategy, which... (one sentence).} 
%See~\cite{abdo2017learning} for a description of and comparison between different strategies. NO NEED TO REFERENCE AGAIN:TIM 
For goal-selection nodes, we sample a child with a probability proportional to that of its goal state based on the goal distribution of the action and template of its parent, as we describe in the next section.

\paragraph{Node Expansion}
\label{sec:node_expansion}
Expanding the selected leaf node (line 7) involves adding its successors of template-selection, goal-selection, and action-selection nodes based on the corresponding actions, templates, and sampled goal states. For the latter, we do so by sampling states $\mathbf{s}_{t+1}$ from the mixture distribution $p(\mathbf{s}_{t+1} \mid \mathbf{s}_t, a,\gamma)$ of the corresponding action $a$ and template $\gamma$, which encodes moving an object $o_k$ relative to one other object $o_l$ specified by the template $\gamma$ (\secref{sec:action_templates}). To restrict the branching factor while exploring several modes of the action, we first draw $S$ poses ${}^{l}\mathbf{T}_k$ from the set of poses of $o_k$ relative to $o_l$ seen in the demonstrations of this action. We use each sample and the pose of $o_l$ in $\mathbf{s}_t$ to compute a corresponding state $\mathbf{s}_{t+1}$ by transforming the pose of $o_k$ to the goal pose ($\mathbf{T}_k(t+1) = \mathbf{T}_l(t)\ {}^{l}\mathbf{T}_k$) as $o_k$ is the only object that moves between $t$ and $t+1$. The resulting samples define the goal distribution, which we discretize by clustering the goal states with respect to $\mathbf{T}_k$ using agglomerative hierarchical clustering such that the state of each cluster is based on the mean pose of $o_k$ over its members. Finally, each cluster is added as a new leaf node to the tree, with a goal probability proportional to the number of samples used to create it. In summary, each node expansion step explores several modes of moving objects relative to each other based on the demonstrations, see \figref{fig:tree}.

\paragraph{Value Initialization and Backpropagation}
\label{sec:backprop}
%\cite{Keller2013TrialBasedHT}
We follow a heuristic-based MCTS approach and gauge how good a newly-added leaf node is as a final goal of the task by initializing its value to the intention likelihood $\Psi(\mathbf{s}_{t+1})$ of its state $\mathbf{s}_{t+1}$. Through backpropagation, we then update all affected node values back to the root such that a node's value depends on those of its children. For this, we use a max-value strategy, which sets the value of a node as the max value of its children while subtracting the constant cost of the action for goal-selection nodes.

\subsubsection{Retrieving the Best Feasible Plan}
\label{sec:retrieve_plan}
%\todo{Add in this paragraph briefly 1-2 sentences about what the plan "step" is that we retrieve when traversing the tree to make the connection to the nodes. Each step is $a_t$, $\gamma_t$, and $\mathbf{s}_{t+1}$ starting from the previous $\mathbf{s}_t$.}
The procedure described in \secref{sec:anytime_search} corresponds to the MCTS-based algorithm we introduced in~\cite{abdo2017learning}. This relies on a lazy approach of checking inverse kinematic constraints and collisions in the state of a leaf node only after it gets selected for expansion to prevent expanding infeasible states. This guarantees the feasibility of the intermediate goal states $\mathbf{s}_{1:T}$, but not of the trajectory between each two states. Additionally, we assumed the existence of a suitable motion planner to generate a trajectory for each step consisting of the current state $\mathbf{s}_t$, the action to be applied $a_t$, the corresponding template $\gamma_t$ and the new desired state $\mathbf{s}_{t+1}$. In this work, we address these limitations by incorporating our action trajectory models when generating the solution plan.

After building the search tree, we traverse it to retrieve the best plan found so far (lines 12-16 of \algref{alg:teach_and_improvise}) while ensuring that the robot can generate and execute feasible action trajectories in each step.
%We repeatedly retrieve the best plan candidate from the tree and check its feasibility until a feasible plan is found.
Procedure $\textsc{RecommendBestPlan}$ greedily traverses the tree from the root to a leaf node while always selecting the child node with the largest value. Each step in the resulting plan corresponds to one level in this path, which represents applying and action $a_t$ in state $\mathbf{s}_t$ using template $\gamma_t$ to move the relevant object and achieve the next state $\mathbf{s}_{t+1}$. To check the feasibility of the resulting plan, we generate the corresponding grasping, moving and releasing trajectories for each action $a_t \in a_{0:T-1}$ using the motion models learned as described in \secref{sec:action_traj}. 
%\todo{explain this part in terms of $a_t$ etc to connect to the plan step in the first paragraph of this section (use $\gamma_t$ of the step instead of $\gamma$ etc). We use bla to set the grasping pose... etc. } 
The grasping trajectory is generated based on the current pose in $\mathbf{s}_t$ of the object associated with action $a_t$ (the grasp pose is learned as part of the model, see \cite{twelsche17iros}), and the motion is performed with respect to the reference frame of the object defined by the corresponding template $\gamma_t$. We set the end pose of each trajectory as the initial state of the next motion model.
%, see also \figref{fig:combined_learning}.
%based on the learned model and the selected template $\gamma$ \todo{link template to reference frame of motion} and goal state for that step. 
%Using the models for grasping, moving and releasing the object and feeding the respective end pose \nichola{which pose?} to the respective \nichola{huh?} successor model, we generate robot gripper and base trajectories for the actions using the approach described in \secref{sec:action_traj}, see also \figref{fig:combined_learning}. 
While generating trajectories, we perform inverse kinematic and collision checks for each pose using $3$D models of the robot and the scene. If any trajectory is found to be infeasible, we update the tree by assigning a value of $-\infty$ to the leaf, backpropagating, and repeating.

\setlength{\tabcolsep}{1pt}
\begin{figure}[t!]
	\centering
	\resizebox{0.8\columnwidth}{!}{%
  	\begin{tabular}{ccc}
  		
 		\includegraphics[width=0.32\columnwidth,trim={12.9cm 4.55cm 14.7cm 1.4cm},clip]{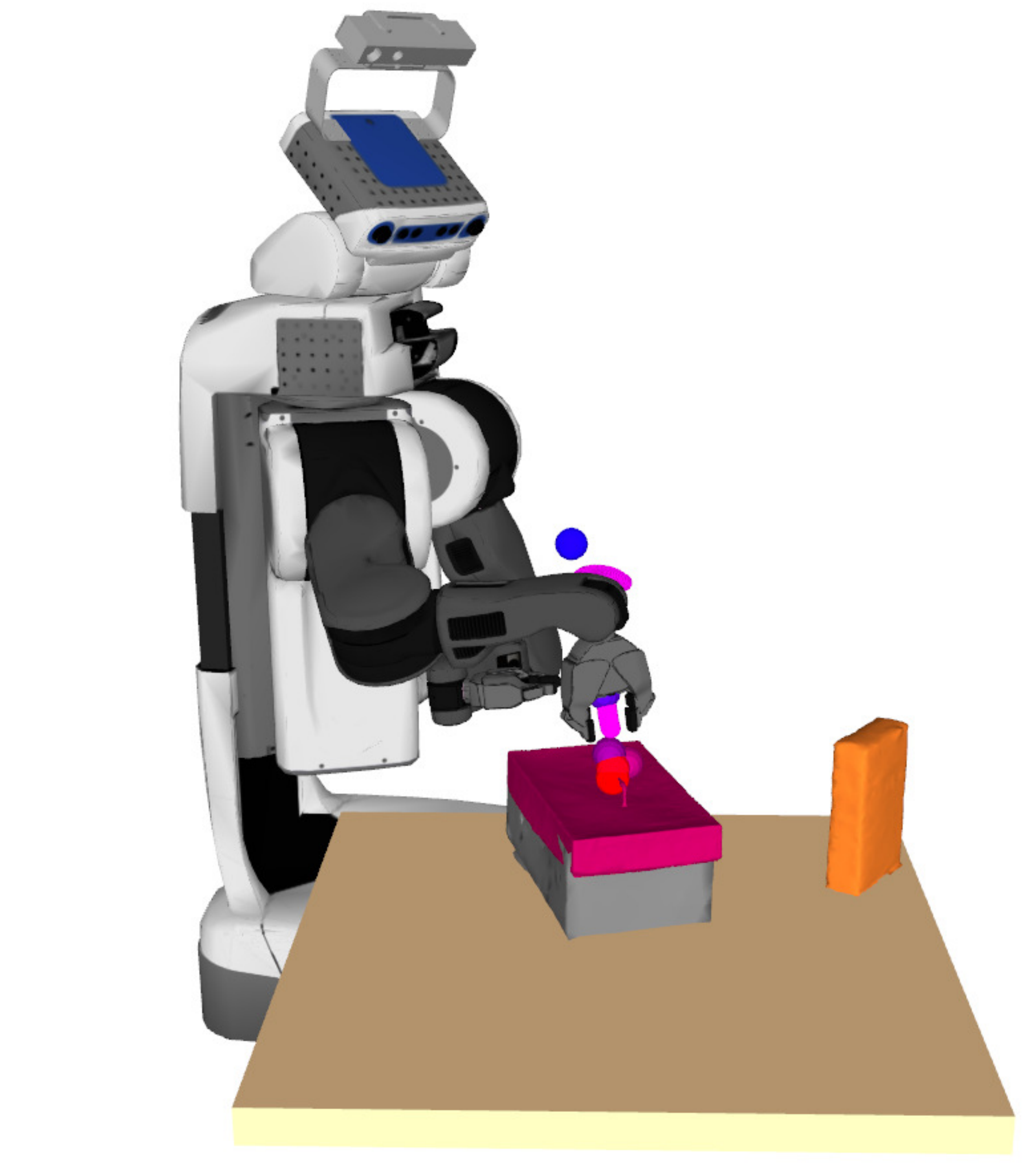} &
 		\includegraphics[width=0.32\columnwidth,trim={12.9cm 4.55cm 14.7cm 1.4cm},clip]{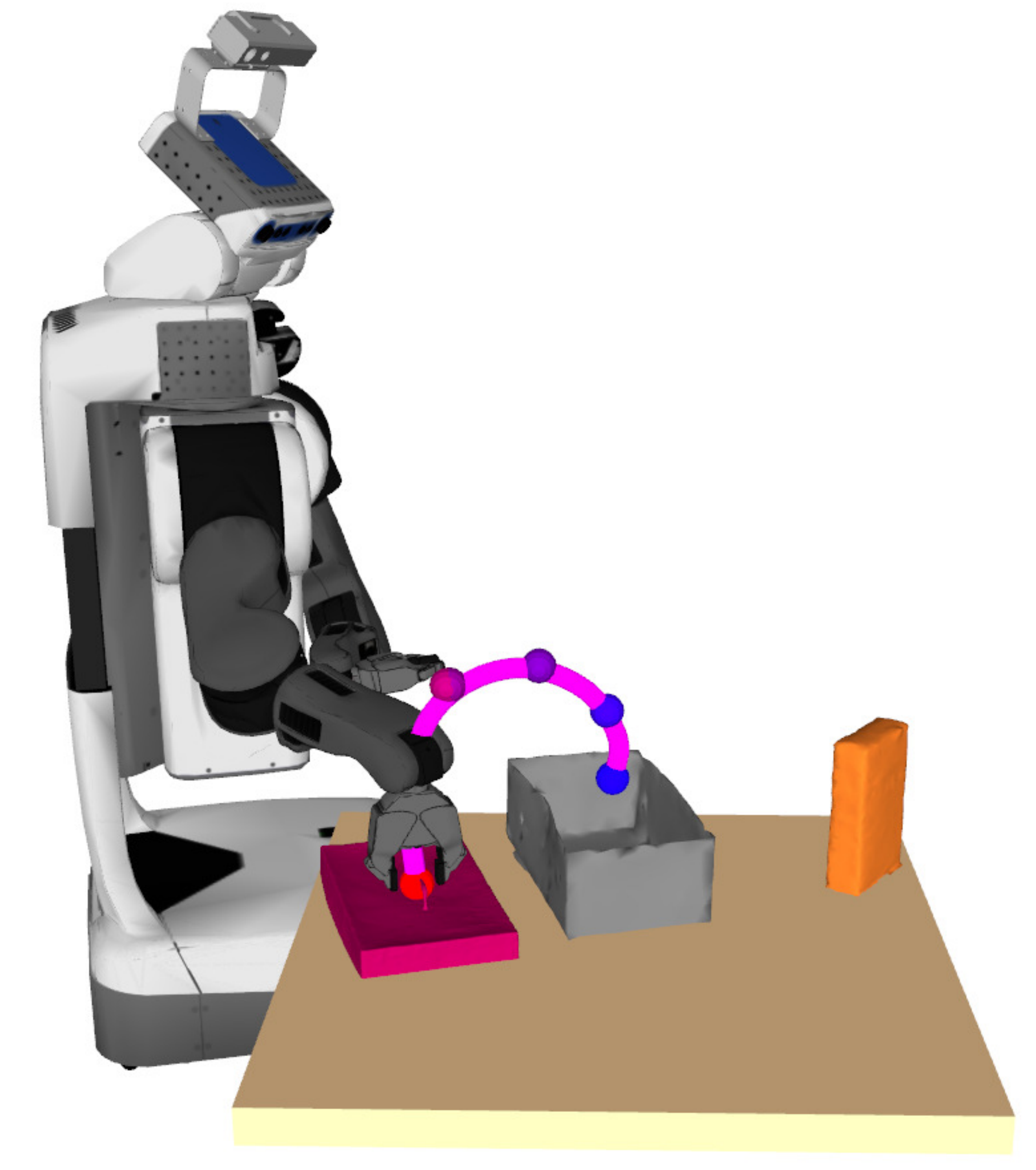} &
 		\includegraphics[width=0.32\columnwidth,trim={12.9cm 4.55cm 14.7cm 1.4cm},clip]{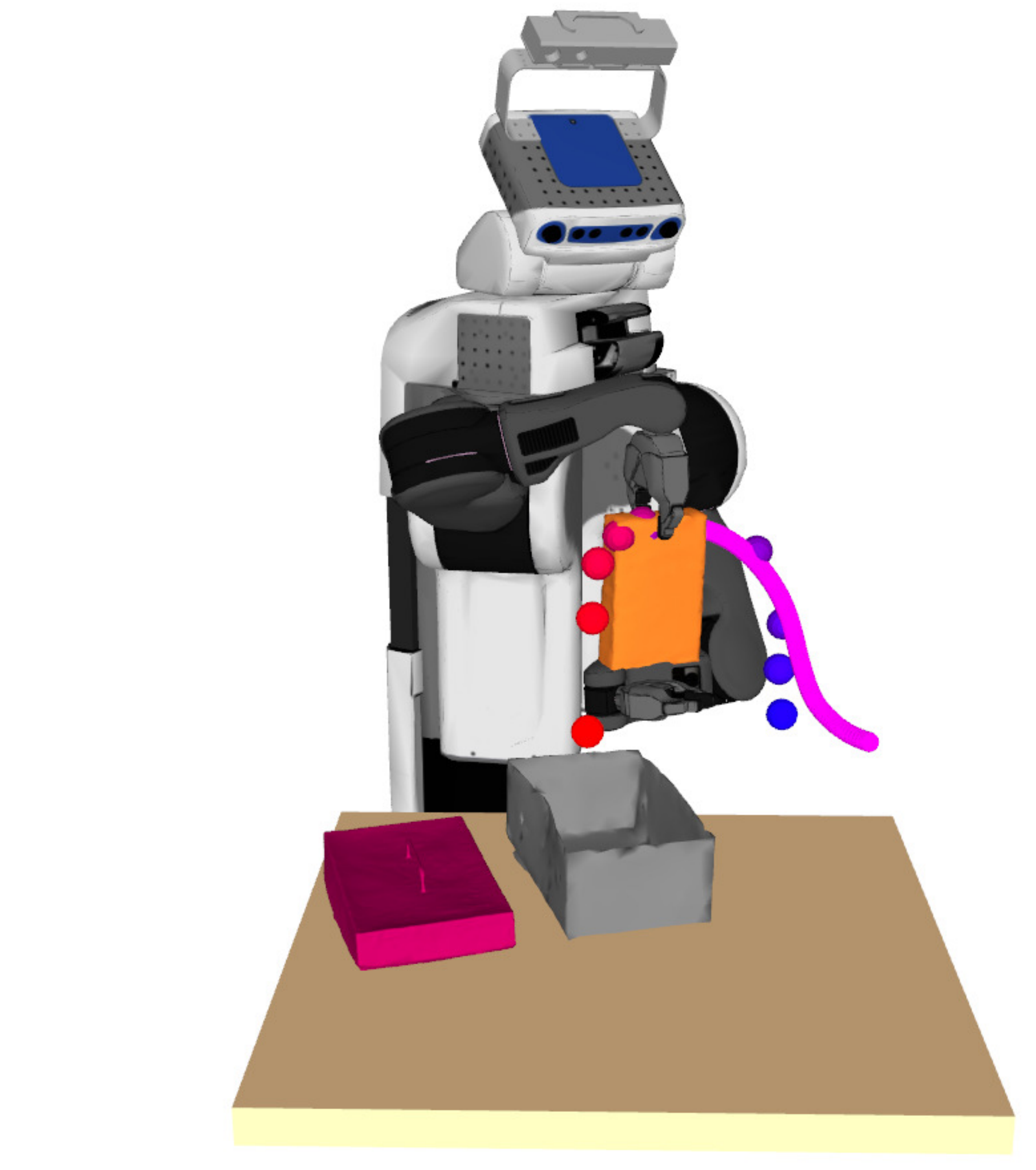}\\
 		\midrule 
 		\includegraphics[width=0.32\columnwidth,trim={11.0cm 5.39cm 18.0cm 0.5cm},clip]{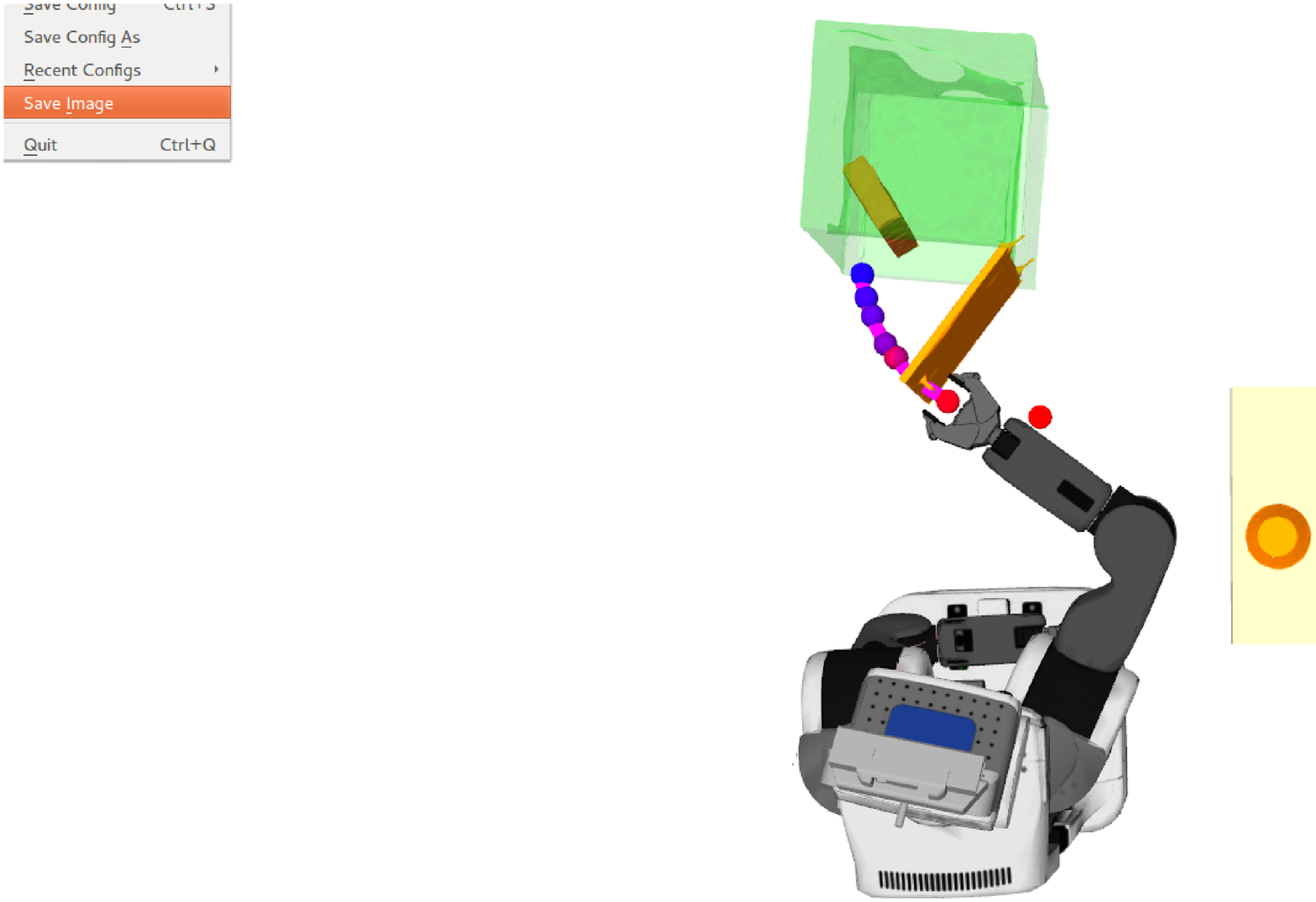} &
 		\includegraphics[width=0.32\columnwidth,trim={11.0cm 5.39cm 18.0cm 0.5cm},clip]{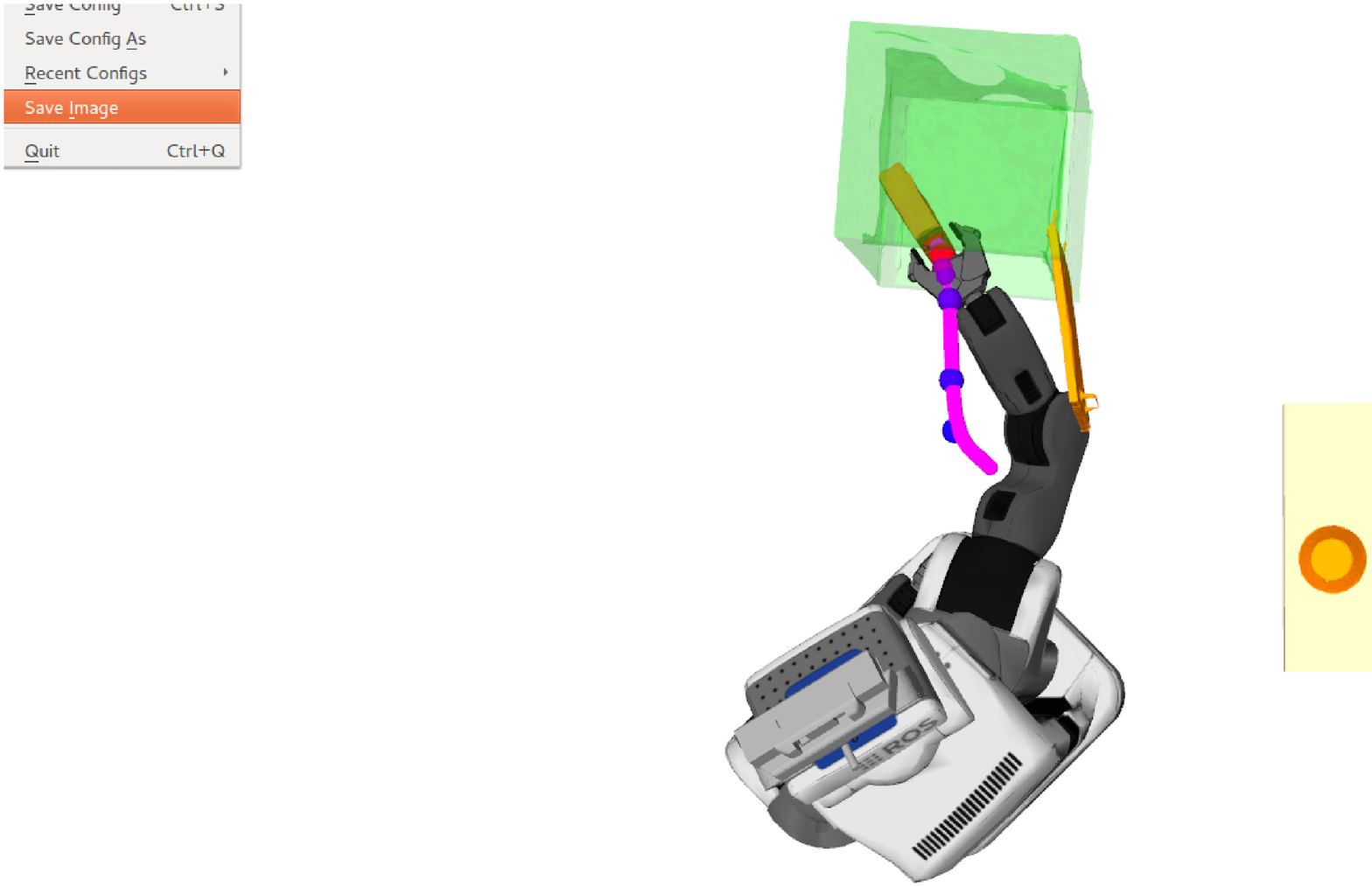} &
 		\includegraphics[width=0.32\columnwidth,trim={11.0cm 5.39cm 18.0cm 0.5cm},clip]{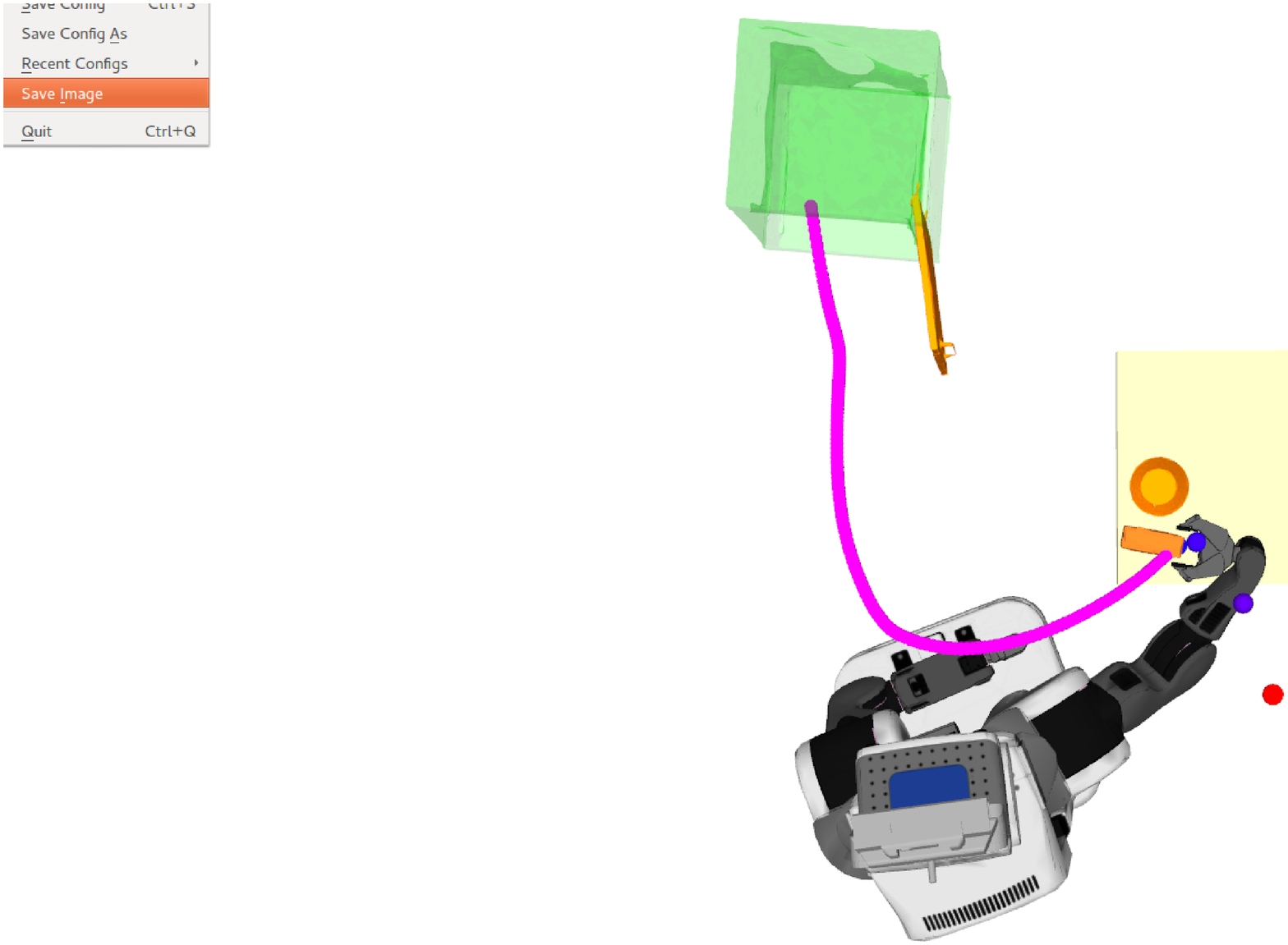}\\
 		\midrule 
 		\includegraphics[width=0.32\columnwidth,trim={9.0cm 2.4cm 6.0cm 0.0cm},clip]{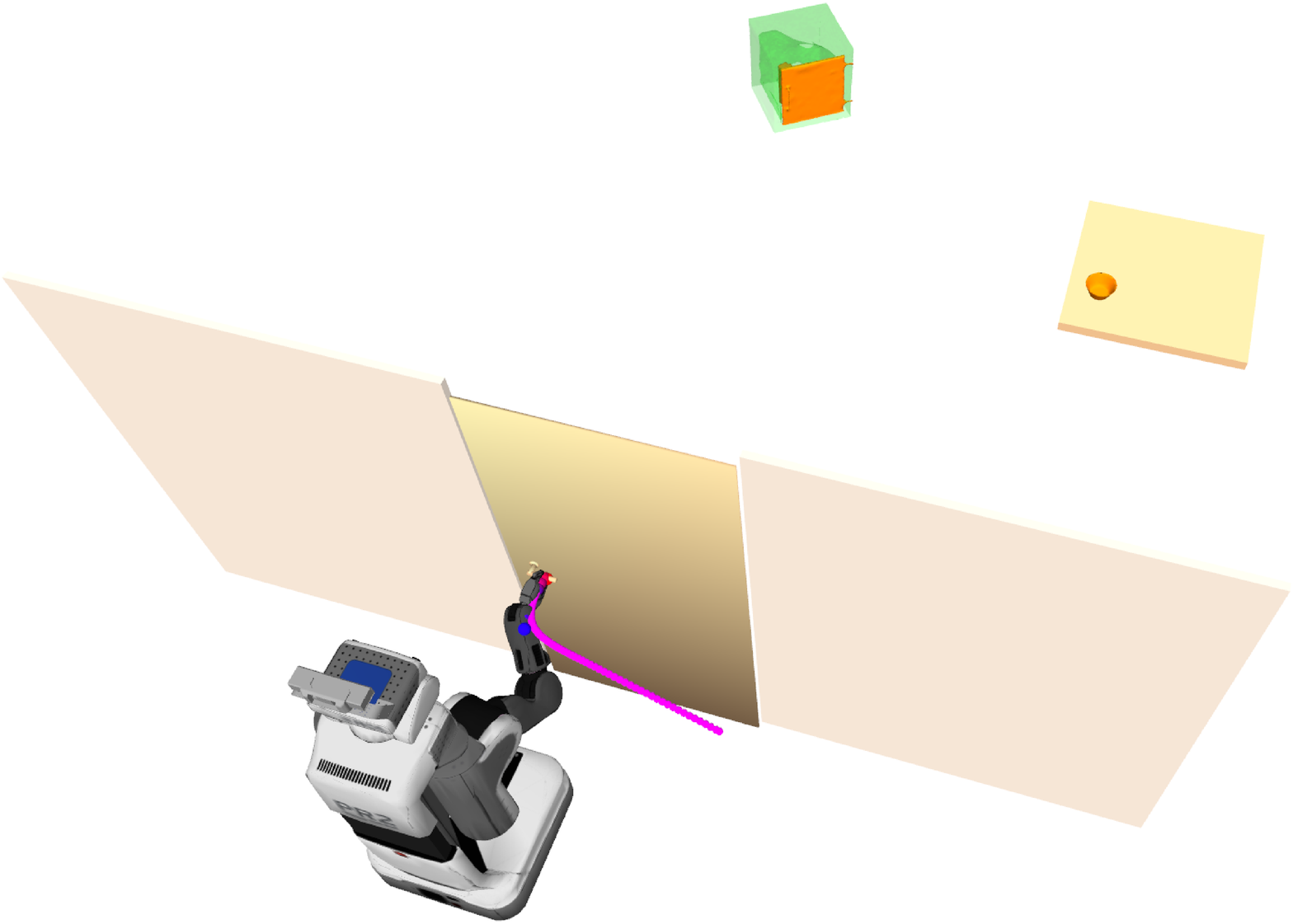} &
 		\includegraphics[width=0.32\columnwidth,trim={9.0cm 2.4cm 6.0cm 0.0cm},clip]{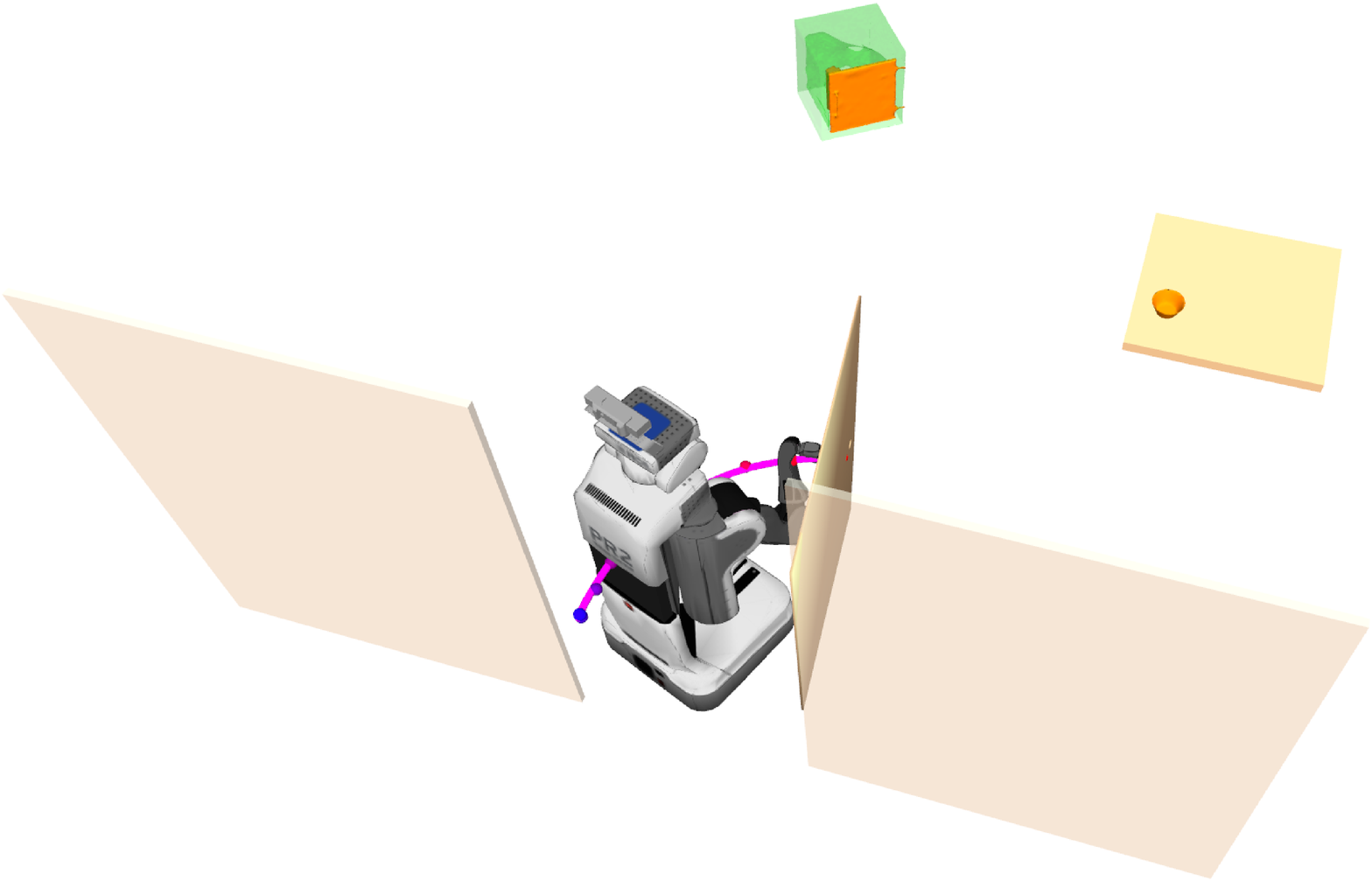} &
 		\includegraphics[width=0.32\columnwidth,trim={9.0cm 2.4cm 6.0cm 0.0cm},clip]{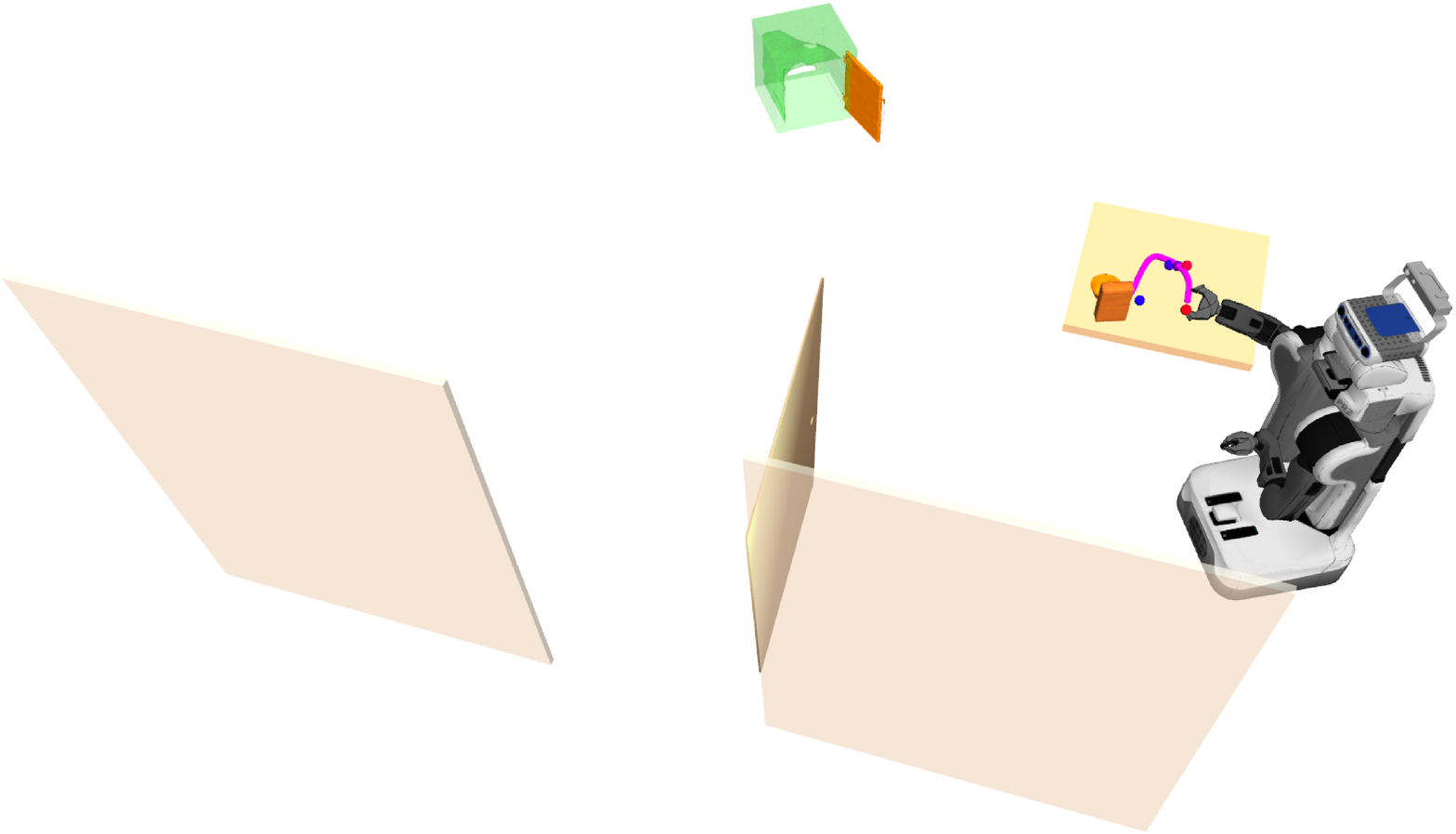}\\
	\end{tabular}
	}
    \caption{This figure shows the three tasks used in the evaluation.
        The first row shows the task of removing the lid of a box and placing
        a second box inside it. In the second task, the robot has to open
        a cabinet door, take out a cereal box and place it next to a bowl
        on the table. In the third task, the robot has to open and move
        through a door before performing the same task as before. The respective action models are shown as colored spheres representing the centers of the Gaussian mixture models used to encode the motion. The generated trajectories for the end-effector are shown in magenta and are used to ensure collision-free solutions that satisfy inverse kinematic constraints when executing each action. %This way feasibility of the plan can be verified.
        %Note that the images display computed robot trajectories and not a simulation of their execution.
        }
    \label{fig:combined_learning}
\end{figure}
\begin{figure}[t!]
	\centering
	\resizebox{0.8\columnwidth}{!}{%
  	\begin{tabular}{ccc}
  		 
 		\includegraphics[width=0.32\columnwidth,trim={2.31cm 0.9cm 1.5cm 0.95cm},clip]{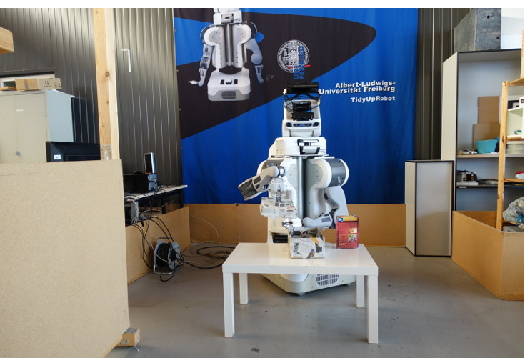} &
 		\includegraphics[width=0.32\columnwidth,trim={2.31cm 0.9cm 1.5cm 0.95cm},clip]{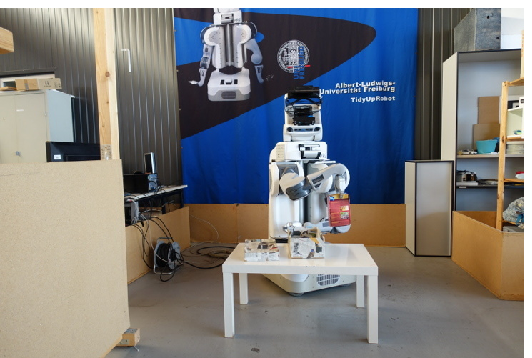} &
 		\includegraphics[width=0.32\columnwidth,trim={2.31cm 0.9cm 1.5cm 0.95cm},clip]{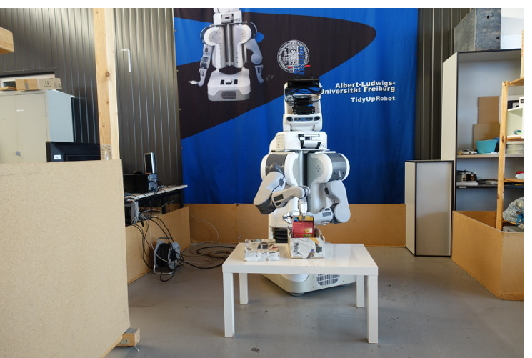}\\
 		\midrule %\hline   \\
 		\includegraphics[width=0.32\columnwidth,trim={4.0cm 0.8cm 1.6cm 1.1cm},clip]{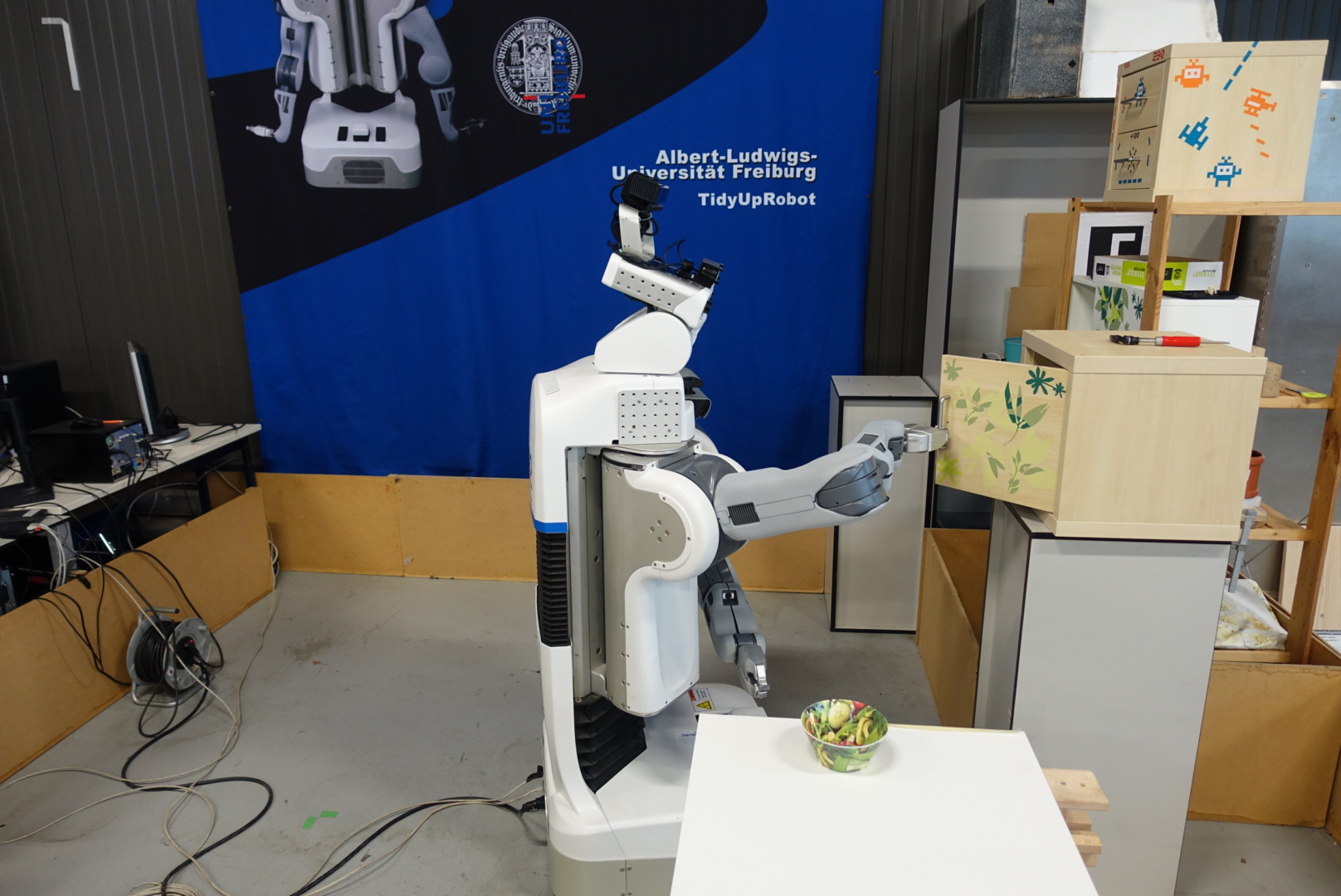} &
 		\includegraphics[width=0.32\columnwidth,trim={4.0cm 0.8cm 1.6cm 1.1cm},clip]{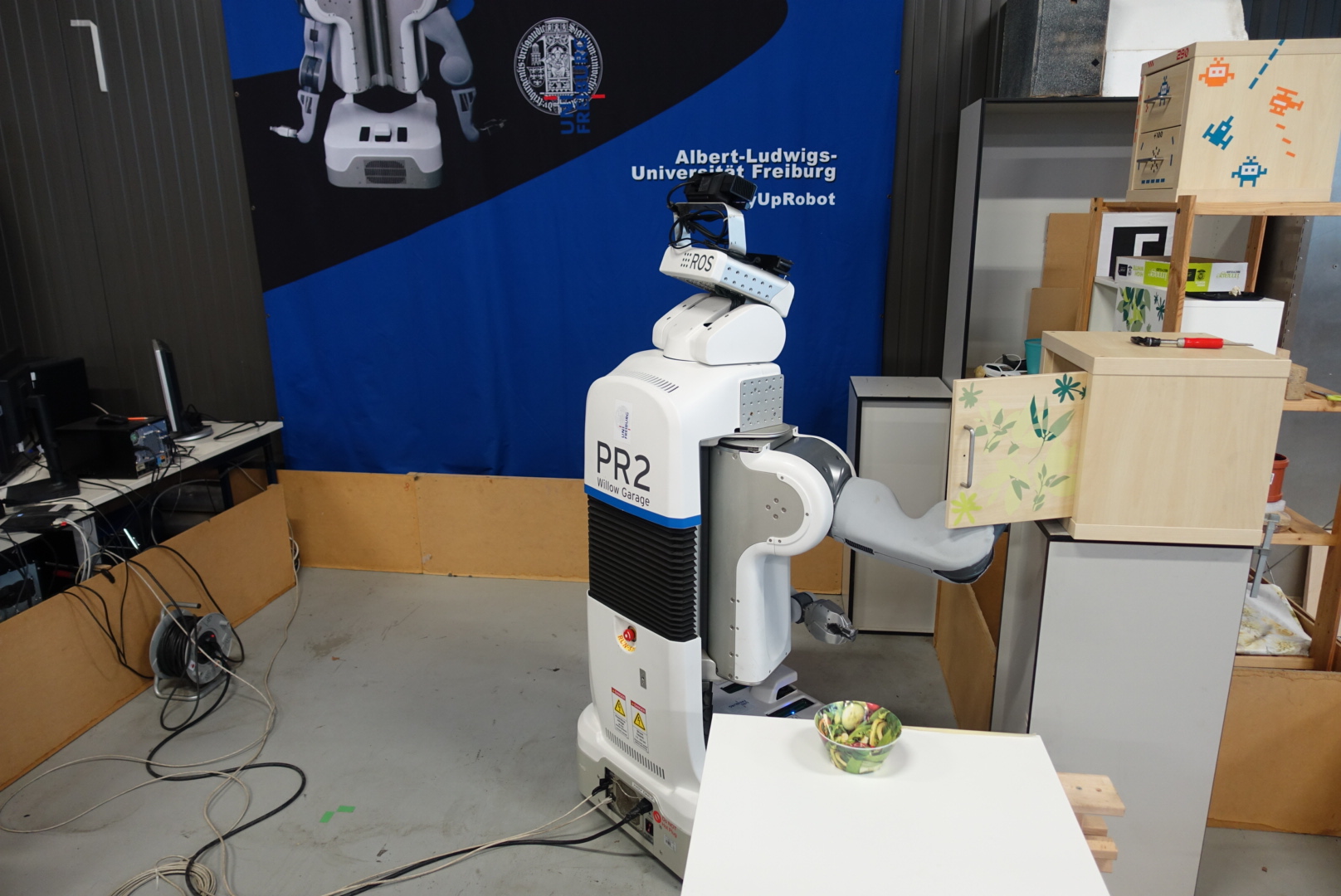} &
 		\includegraphics[width=0.32\columnwidth,trim={4.0cm 0.8cm 1.6cm 1.1cm},clip]{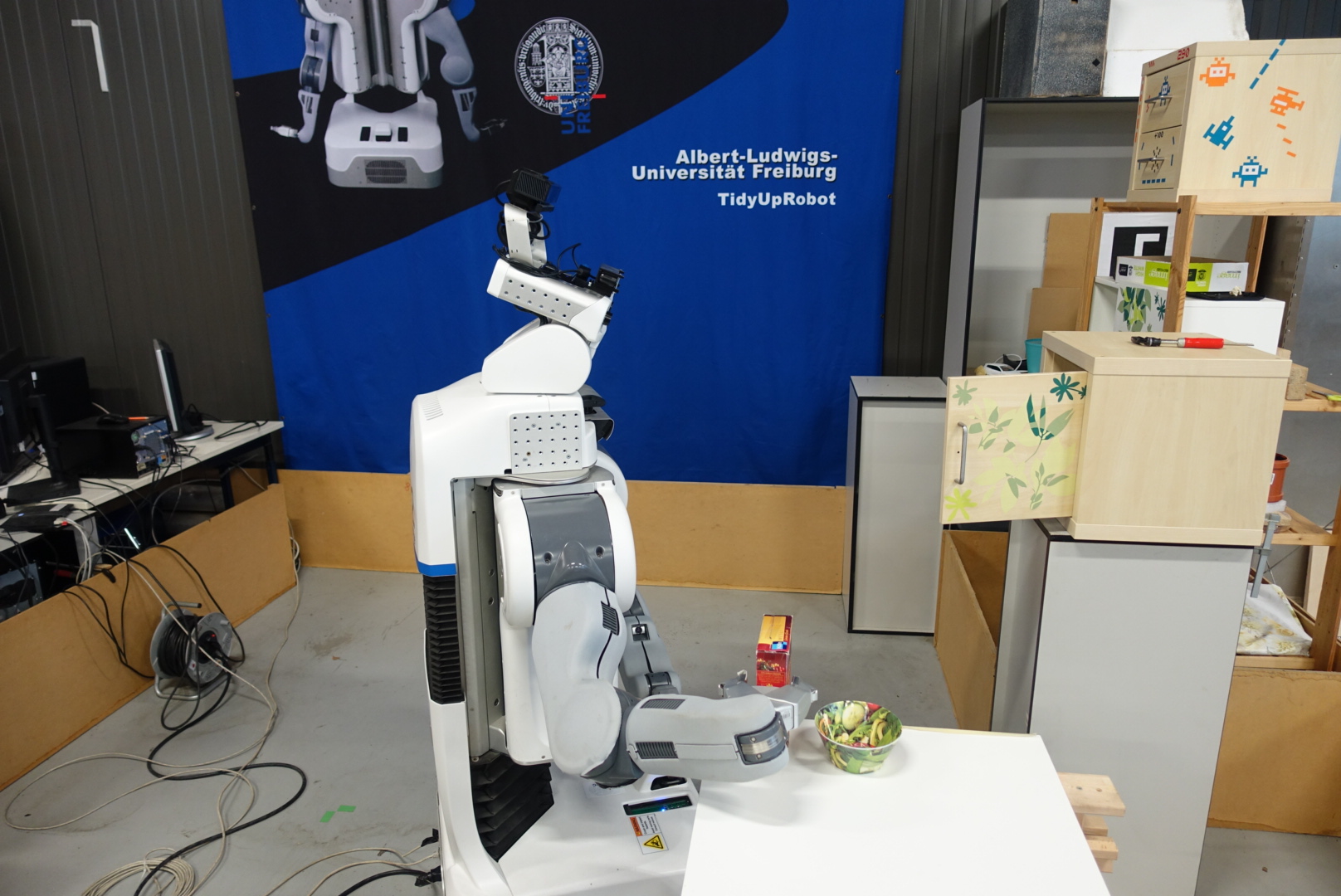}\\
 		\midrule 
 		\includegraphics[width=0.32\columnwidth,trim={0.1cm 0.1cm 1.0cm 0.5cm},clip]{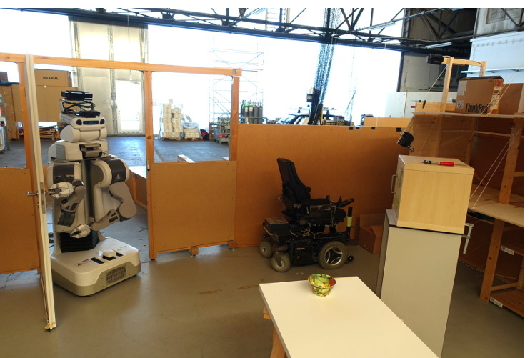} &
 		\includegraphics[width=0.32\columnwidth,trim={.1cm 0.1cm 1.0cm 0.5cm},clip]{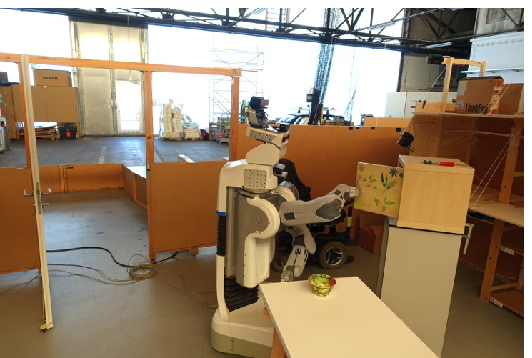} &
 		\includegraphics[width=0.32\columnwidth,trim={.3cm 0.1cm 0.8cm 0.5cm},clip]{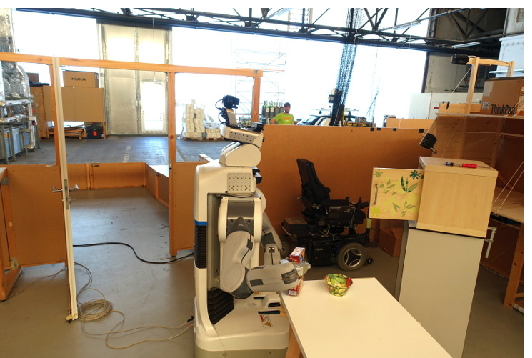}
	\end{tabular}
	}
    \caption{Examples of the robot executing the sequences of actions generated by our approach for solving the tasks in~\figref{fig:combined_learning}. Our approach allows the robot to generate feasible trajectories for each step of the plan using the learned action models and without requiring an existing motion planner.}
  	\label{fig:robot_experiments}
\end{figure}

\setlength{\tabcolsep}{6pt}

\section{Experimental Evaluation}
\label{sec:experimental_setup}

%We evaluate our approach in simulated and real-world robot experiments. 
%As our system is based on previous works for task imitation and action learning that have been thoroughly evaluated \nichola{we can't say it like that. I need to rephrase after sleeping.} 

In this section, we present the experimental evaluation of our proposed approach in both simulation and with a real robot. To record teacher demonstrations, we rely on \cite{zimmermann18icra} to track the hand of the teacher using RGB-D images, and on \textit{Simtrack}~\cite{pauwels15iros} to detect the objects in the scene and estimate their 6-dof poses. 
We segment the demonstrations automatically based on which object is being manipulated as described in \secref{sec:problem}. 
We use our approach (\secref{sec:action_learning}) to learn the motion model of each action.
% by adapting the observed teacher trajectories to the capabilities of the robot. We use the intermediate recorded states to learn the action goal distributions as in \secref{sec:action_templates}. 
Additionally, we use the final state after each task demonstration to learn a model of the intention likelihood of the task as in \secref{sec:intention}. Finally, we rely on the \textit{MoveIt!} library \footnote{[Online] Available:\url{http://moveit.ros.org}} %\cite{moveit} 
to perform collision and inverse-kinematics checks.%, and on \todo{bla} to execute generated trajectories. 
%\todo{What about localizing in the room after entering the door and moving from the door to the cabinet? List any assumptions here like a map or localization library. TIM: True. Had it, removed it due to space... Added it again in B but not sure if essential...}

Our previous work and extensive evaluation focused on either learning complex motion models of individual actions \cite{twelsche17iros}, or on generalizing sequential tasks consisting of simple point-to-point, pick-and-place actions that can be executed using standard motion planners \cite{abdo2017learning}. 
In \cite{abdo2017learning}, 
we also discuss computation cost and search efficiency for different variants of our teach-and-improvise algorithm.
%we also discuss computation cost and search efficiency of the planning.  
Therefore, in this work, we conducted experiments to demonstrate the benefits that our proposed integrated system brings: \emph{i)} the ability to jointly learn task and action models from a few demonstrations; \emph{ii)} the ability to imitate tasks involving geometrically-constrained manipulation actions without requiring an existing motion planner that depends on articulated object models (e.g., a model of a door); and \emph{iii)} the ability to improvise task solutions in previously unseen settings. We investigate three tasks of increasing complexity, see
\figref{fig:combined_learning}.

\subsubsection*{Task 1}
In this task, the teacher demonstrated grasping the lid of a box, opening the box, placing the lid next to the box, grasping a second box, and placing it inside the opened box (see \figref{fig:intro}). We provided the robot with five demonstrations to learn both the task goal and the involved action models.

\subsubsection*{Task 2}
The second task consists of opening a cabinet door, taking a cereal
box from the cabinet, and placing it next to a bowl on the table. Note that our proposed approach enables the combined learning of the actions and the task intention likelihood from the same set of task demonstrations as in \textit{Task 1}. However, in this work, we do not explicitly address the perception problem. For this task, we provided the robot with demonstrations of the actions independently of the demonstrations of the final task state, as the scene is often not fully observable from a single perspective. Overall, we provided five demonstrations of the task goals and for each action. Beyond the recorded state during the demonstrations, we did not provide the robot with further information such as a model of the articulated cabinet door.
%For this task the overall goal and the low level part of the action models are learned using independent task and action demonstrations. We use $5$ demonstrations of the overall task to learn the task goals as well as each $5$ demonstrations of the individual actions to learn the respective action models. The task is designed to show that our approach is able to imitate complex task including geometrically constrained actions like opening the cabinet without inferring further information about the world.

\subsubsection*{Task 3}
This task involves the same goal as in the previous task. Here however, the robot is initially outside of the room and has to first enter through a door to reach the cabinet and the table. For this task, we use the demonstrations from \textit{Task 2}, and additionally provide the robot with $10$ teacher demonstrations of the door opening action only.

\subsection{Generalizing the Learned Tasks in New Settings}
\label{sec:combined_learning}

To demonstrate the ability of our system to learn applicable models of the task and actions and use them to generate plans starting from new states $\mathbf{s}_0$, we evaluated our approach on \textit{Task 1} and \textit{Task 2} each for $50$ different simulated settings. 
The starting poses of the objects were sampled from the space of reachable poses, e.g., pose of the box inside the cabinet is geometrically constrained by robot's grasping capabilities.
Besides the learned models no additional information or explicit goal state was provided. In this experiment the generated action sequences are checked for feasibility, the execution of the generated plans is addressed in \secref{sec:experimental_evaluation}.
%Quantitatively, we generated plans for solving \textit{Task 1} and \textit{Task 2} each for $50$ different settings.
%\todo{define settings. It should be explicit like 50 random starting states. TIM: the starting states are not random but manually chosen. Random would only work within certain bounds due to the restrictions posed by the action models, especially grasping the box in the cabinet .}

For \textit{Task 1} our approach generated feasible solutions for $43$ trials and for \textit{Task 2} in $41$ trials.  Note that the computed plan was not always the same as our approach explores different action sequences and interpretations, e.g., the robot may choose to move the lid relative to itself or to the box to solve the task.
Failures occurred due to the infeasibility of the required actions, i.e., there was no collision-free plan. 
Out of the seven failures for \textit{Task 1}, two still resulted in a partial solution of removing the lid of the box. 
For \textit{Task 2}, the nine failures included six results with a partial solution. When, for instance, the grasping of the box inside the cabinet with the learned action model is not possible due to collisions, our approach proposed to open the cabinet and stop afterwards as this is the best achievable solution. This highlights the advantage of our approach in improvising solutions by maximizing the intention likelihood of the task.

%\todo{I would say depending on what is more feasible or something like that, not "yielding the same result." Also, this example is about the template chosen to execute the action, not the sequence or type of the actions. It would be better if there's an example with a different sequence altogether. TIM: The order cannot change due to the obvious geometric restrictions. What changes are the templates used...}
%\todo{You need to define what "success" and "failure" is for each task. TIM: Would definitely be nice but since every line added means another line needs to be removed... I think it is obvious what success, failure and partial solutions are here.}

%While the large number of successful task imitations implies that our systems has learned the intention of the task, this is even visible in the partial solutions as here the robot tries to fulfill as much of the task as it can given its capabilities.

\subsection{Task Imitation on the Robot}
\label{sec:experimental_evaluation}
%\todo{Same points as in the previous section regarding organizing the thoughts and what messages to highlight.}
To demonstrate that the plans generated by our approach are executable in practice, we evaluated the tasks on our PR2 robot in real-world and simulation experiments. 
%\nichola{again, confused. So the previous experiments didn't involve a robot model? TIM: They did involve a robot model. But we just generated robot trajectories, that is not the same as actually executing them in simulation (eg. with gazebo).}
%After learning the task and the underlying actions our system takes the poses of all involved objects in the scene and generates a plan consisting of a sequence of feasible manipulation actions to apply, including their respective goal poses.
The three rows in \Cref{fig:combined_learning,fig:robot_experiments} show examples of the planning process and the execution of the three tasks, respectively.  For
\textit{Task 1}, the object poses were detected before generating the
plan using \textit{Simtrack}~\cite{pauwels15iros}. We reproduced the task from
five different starting states. As no navigation with the mobile base was required here, we used a fixed robot position to minimize the errors induced from object detection
and localization. 
\textit{Task 2} was planned and executed in five trials in both simulated and real-world settings, each with different starting poses of the objects and the robot.
%\nichola{(mentioning simulation here is weird. The previous section should be about simulation and this one about the real robot). TIM: The previous section was about generating the plans, not executing them. This section is about showing that the generated plans are executable. For this we use both real-world and simulation experiments. I guess this misunderstanding made the section a bit weird when you rephrased them last time :)} 
For \textit{Task 3}, we performed three repetitions both in simulation and real-world settings. For \textit{Tasks 2} and \textit{3}, we manually provided the initial object poses to the robot as some of them were initially occluded, e.g., behind the respective doors.

%We relied on an existing map of the scene for the robot to localize itself during task execution.

For all tasks, our approach was able to compute feasible solutions from each initial state and to successfully execute them on the robot. This demonstrates that our approach enables transferring human demonstrations to the robot such that it can reproduce them based on its own constraints. In \textit{Task 3}, our algorithm always selected action sequences that require the robot to first apply the door opening action as all other solutions resulted in collisions with the door (see attached video). The robot is able to achieve this by generating feasible trajectories from the learned action models without requiring a motion planner as in~\cite{abdo2017learning}, i.e., our approach enables incorporating such constrained mobile manipulation actions when learning from non-expert teachers. Furthermore, the imitation performed in \textit{Task 3} highlights the flexibility of our approach in incorporating actions learned in a different context to solve a different task.
%our approach enables learning from demonstrations that transfers directly from humans to a robot for complete tasks. 

%This shows that our approach is able to adapt to new scenarios while pursuing the demonstrated goal and improvise executable plans . 

\section{Conclusion}
\label{sec:discussion}

In this paper, we presented a novel approach to learning sequential mobile manipulation tasks demonstrated by a human teacher. Our work combines learning action models that allow reasoning about feasibility on a geometric level with a probabilistic planning framework based on Monte Carlo tree search. Our approach adopts flexible, probabilistic models of each action and the overall goal of the task based on the spatial relations between the involved objects. This allows the robot to reason about different valid ways of imitating the task, thus tackling the ambiguity in the demonstrations. 
As opposed to standard planning approaches that can leverage this prior knowledge, we formulate solving the task as an optimization problem to maximize the intention likelihood given the demonstrations, thus avoiding the need for the teacher or an expert to specify an explicit goal state in each situation, i.e., the robot would \emph{do what it can} to generalize the demonstrations given a new initial state. 
Furthermore, our approach enables the robot to reproduce complex, geometrically-constrained action trajectories and to automatically select a suitable reference frame for each motion without the need for a motion planner that requires prior semantic knowledge of the task and its constraints. This makes our approach unique in its ability to imitate sequential manipulation tasks demonstrated by a non-expert teacher without a
%in cases where there is no available 
formal planning domain description of the task or expert knowledge about the relevant reference frames for each action. 
In experiments we demonstrate the effectiveness of our approach in computing and executing feasible sequences of actions to solve complex mobile manipulation tasks and generalize them in new settings.

\bibliographystyle{IEEEtran}
\footnotesize
\bibliography{refs,grounding_modules}

\end{document}